%% file: acl_latex.tex
\pdfoutput=1
\documentclass[11pt]{article}

\usepackage[final]{acl}

\usepackage{times}
\usepackage{latexsym}

\usepackage[T1]{fontenc}
\usepackage[utf8]{inputenc}

\usepackage{microtype}

\usepackage{inconsolata}

\usepackage{graphicx}

\usepackage{booktabs}
\usepackage{makecell}
\usepackage{hyperref}

\usepackage{booktabs}
\usepackage{multirow}
\usepackage{xspace}

\usepackage{multirow}
\usepackage{amsmath} % for "\sideset" macro
\title{Leveraging Large Models to Evaluate Novel Content \\ A Case Study on Advertisement Creativity}

\author{
    Zhaoyi Joey Hou, Adriana Kovashka, Xiang Lorraine Li 
    \\ Department of Computer Science
    \\ University of Pittsburgh 
    \\ \texttt{\{joey.hou, kovashka, xianglli\}@pitt.edu}
}

\begin{document}
\maketitle

\input{sections/0_abstract}
\input{sections/1_intro}

\input{sections/2_related_work}
\input{sections/3_dataset}

\input{sections/4_exp}

\input{sections/5_results}
\input{sections/6_conclusion}

\bibliography{custom}

\input{sections/appendix}

\end{document}

%% file: sections/0_abstract.tex
\begin{abstract}
Evaluating creativity is challenging, even for humans, not only because of its subjectivity but also because it involves complex cognitive processes. Inspired by work in marketing, we attempt to break down visual advertisement creativity into atypicality and originality. With fine-grained human annotations on these dimensions, we propose a suite of tasks specifically for such a subjective problem. We also evaluate the alignment between state-of-the-art (SoTA) vision language models (VLMs) and humans on our proposed benchmark, demonstrating both the promises and challenges of using VLMs for automatic creativity assessment\footnote{\url{https://github.com/JoeyHou/ads-creativity}}.
\end{abstract}

%% file: sections/1_intro.tex
\section{Introduction}

Creativity, the ability to generate novel and high-quality ideas, is a fundamental human cognitive ability. Research in a variety of domains has been devoted to understanding creativity, including psychology focusing on human creativity~\cite{Olson2021-xq, Guilford2012-kh, Alabbasi2022-jz}, machine learning on machine creativity~\cite{Ismayilzada2024-iq, creativity_survey_deep_learning}, human-computer interaction (HCI) on the combination of both~\cite{lu2024aihumanityssalieriquantifying, Chakrabarty_art_artifice, Porter2024-pp, Marco2024-pc}, and marketing on advertisement creativity~\cite{modeling_determinants, whatdoweknow, HowAdvertisingCreativityWorks}. 

\begin{figure}[t!]
    \centering
        \includegraphics[width=0.48\textwidth]{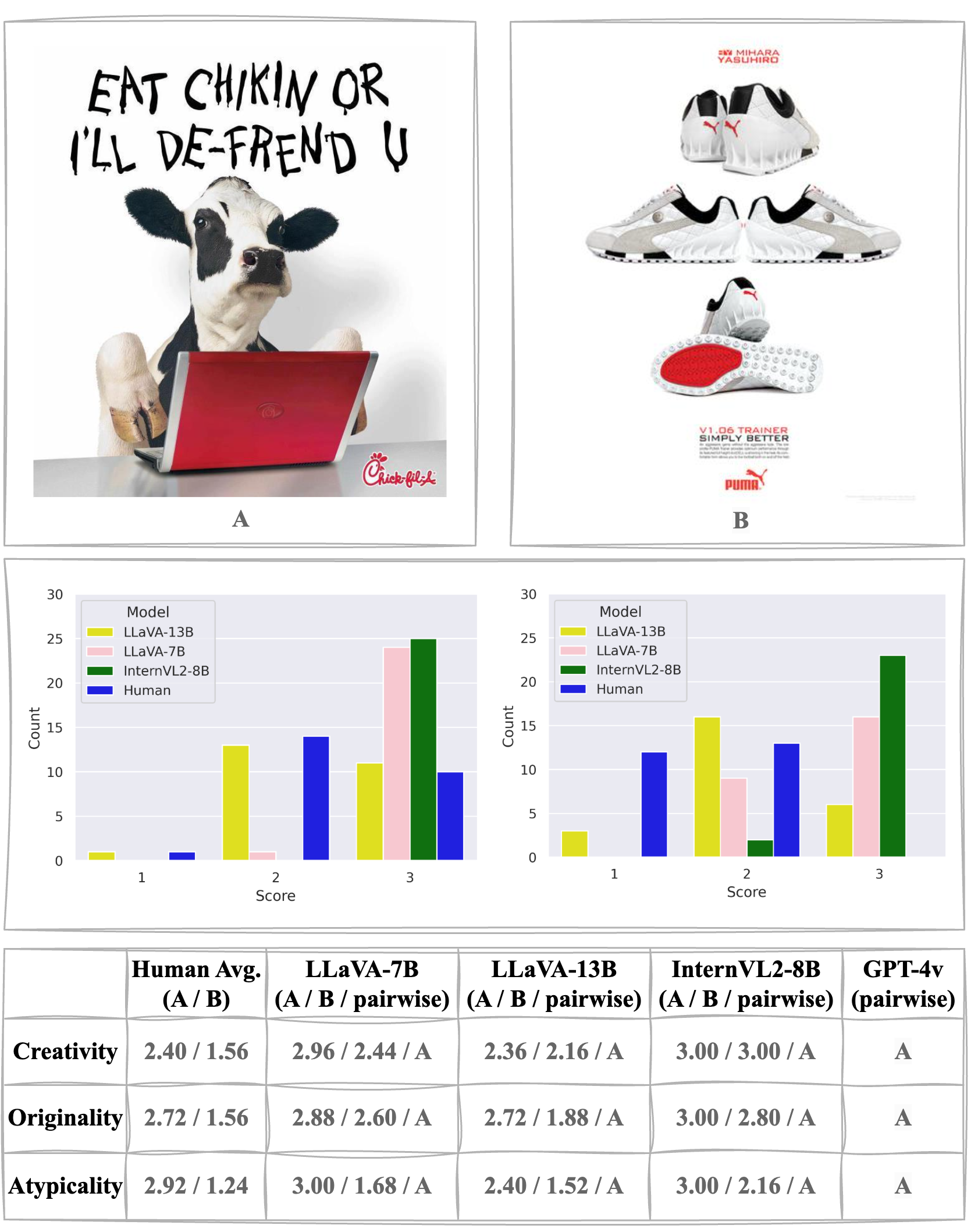}
    \caption{Top: two advertisements from the dataset; Middle: human and VLM rating distribution on creativity (25 each); Bottom: average human and VLM ratings, and VLM pairwise predictions on creativity, originality, and typicality; Scores are on a scale of 1 to 3, 3 being the best. Both humans and models agree that advertisement \textbf{A} is more creative than \textbf{B}; however, their score distributions differ substantially, indicating the importance of measuring it.
    }
    \label{fig:intro}
    \vspace{-7mm}
\end{figure}

In particular, in the marketing domain, studies have shown positive effects of creative advertisements on consumer behavior, including increased purchase intent and positive brand impression ~\cite{sharma2012advertising, terkan2014importance}. Therefore, advertisement creators are motivated to develop and evaluate creative advertisement content consistently. Extensive research has been conducted to understand what the general public would consider as creative when it comes to advertisements ~\cite{El-Murad188, HowAdvertisingCreativityWorks, doi:10.1080/13527266.2012.677464, modeling_determinants}, many of which consider advertisement creativity to be a combination of ``atypicality'' (or, synthesis, abnormality, etc.) and ``originality'' (or, novelty, uniqueness, etc.). Advertisement (A) in Figure~\ref{fig:intro} is an example that is both atypical and original. It is atypical because cows do not usually type on a computer. The text \textit{``Eat chikin or I'll de-friend U''} (which consists of multiple slang expressions), the cow, the computer, and the small logo of Chick-fil-A are a rarely-seen combination, given that it is a fast-food advertisement, making it very original. Decoding advertisement creativity with such a complex combination of concepts and ideas requires extensive reasoning, making the evaluation of creativity a challenging task. Unsurprisingly, previous work \cite{Chakrabarty_art_artifice, tian-etal-2024-macgyver} has heavily relied on human evaluation, but human annotators are expensive and often inaccessible.

Recently, foundational models demonstrate impressive performances in other evaluation tasks, such as summarization ~\cite{zhong-etal-2022-towards}, Long-Form QA~\cite{Jiang2023TIGERScoreTB}, and commonsense text generation~\cite{xu2023instructscore}, many of which were previously dominated by humans. For creativity evaluation, prior work \cite{Chakrabarty_art_artifice, Atmakuru2024-th, Ismayilzada2024-ug} explored the ability of large language models (LLMs) to assess writing creativity. This poses the question of whether we can leverage foundational models to expand automatic evaluation of creativity to multi-modal data, such as visual advertisements, with the help of VLMs.

To this end, we conduct several fine-grained, automatic creativity evaluations for visual advertisements, including creativity, originality, and atypicality. We decompose creativity into atypicality and originality and then collect high-quality human ratings of advertisements in those dimensions, as shown in Figure \ref{fig:intro}. We experiment with SoTA VLMs to predict these ratings and examine the human-model alignment in both intrinsic (i.e., one image at a time) and pairwise fashion. In contrast to the traditional emphasis on prediction accuracy, we evaluate models' ability to capture the task's subjective nature and to gauge annotator disagreements. We also show that VLMs perform impressively in a pairwise setting, reasonably well in distribution modeling, and less effectively in disagreement. Our results highlight the importance of both high-quality annotations and distribution modeling for subjective tasks. We believe our benchmark and evaluation metrics provide a solid foundation for utilizing VLMs to assist visual content creators.

%% file: sections/2_related_work.tex
\section{Related Work}
\textbf{Evaluation of Creativity} 
Research in the evaluation of creativity includes cognitive science \cite{SaidMetwaly, QuantifyingcreativityDeanKeithSimonton, lloydcox2022}, marketing \cite{El-Murad188, HowAdvertisingCreativityWorks, doi:10.1080/13527266.2012.677464, modeling_determinants}, creative writing \cite{LiquidGoldDowntheDrainMeasuringPerceptionsofCreativityAssociatedwithFigurativeLanguageandPlay}, human computer interaction \cite{chakrabarty2024creativity}, and artificial intelligence \cite{chakrabarty-etal-2023-spy, Chakrabarty_art_artifice, Atmakuru2024-th, Ismayilzada2024-ug}. %Although specific focuses vary, 
There are two consensuses among those works: First, creativity requires both effectiveness (i.e., clearly conveying a message or solving a problem) and divergence (i.e., diverging from the norm or commonly seen objects). We ensure effectiveness via quality check questions in human annotation and mainly focus on modeling divergence with atypicality and originality. Second, the evaluation of creativity is subjective.
This motivates our distribution modeling task instead of a majority-label prediction task.
Our work is closely related to \citet{modeling_determinants}, which focused on advertisement images and proposed five creativity dimensions, including atypicality and originality. We adopt their creativity decomposition.

\textbf{Automatic Evaluation of Creativity with Foundation Models} 
There has been much recent work that automatically evaluates the creativity of given content with foundational models. For example, \citet{assessing_creativity_llms_math, neocoder} evaluate creativity in problem-solving tasks (i.e., math and coding); \citet{Chakrabarty_art_artifice, Ismayilzada2024-ug, Lu2024-ma, Atmakuru2024-th} focus on creative writing tasks; \citet{Organisciak2023-pb, push_gpt_creativity_aut, Chen2023-tb, Bellemare-Pepin2024-yc, assessing_understanding_creativity_llm} study how to apply psychological tests to evaluate LLM creativity; and \citet{Yiming2025-xs, McLaughlin2024-jb} study creative question answering. All of these focus on text-only data, while we focus on visual content. One related work that also works with multi-modal data is \citet{zhong2024let}, where the focus is on creative humor generation, instead of creative advertisement understanding.

%% file: sections/3_dataset.tex
\section{Dataset}

\input{sections/table_questions}

% \subsection{Ads Dataset}

We use the Pitt Advertisement Dataset (referred to as \texttt{Pitt-Ads}) as our basis, where each advertisement image is annotated with its topic (10 topic groups in total), and expected actions from viewers after seeing the advertisement. Additionally, the advertisements are annotated with the category of atypical objects in it \cite{Hussain2017AutomaticUO, ye2019interpreting}\footnote{Defined by the original authors of the dataset as ``non-photorealistic'' objects within an advertisement, e.g., a farm inside of a cup of ice cream, demonstrating the freshness of milk in the ice cream (Figure \ref{fig:ad_example_atypicality}).
% \href{https://joeyhou.github.io/images/ad_example_atypicality.png}{(click to see)}. 
An advertisement can feature any number of atypical objects or none at all. }.  We sample 100 advertisements and collect fine-grained human creativity annotations (\texttt{Creative-100}). We also sample an additional 300 advertisements from the remaining data points for atypicality prediction (\texttt{Atypical-300}).

\subsection{\texttt{Creative-100}}
% \lorraine{We should start this section by briefly introduce Creative-100 data. What are the sample topics? why do we want to do this task collecting multiple responses?}
\label{sec:fine_grained_creativity_data}
\texttt{Creative-100} consists of 100 advertisements, with 10 from each topic group: food, drinks, clothing, non-commercial, automobile, beauty, service, leisure, electronic, and healthcare. For fine-grained creativity evaluation, we include two dimensions of creativity: originality and atypicality, which are the top two indicative aspects of advertisement creativity according to \citet{modeling_determinants} (more details in Appendix \ref{sec:app-selected-dimensions}). These two dimensions (originality, atypicality) and the overall creativity combined are the fine-grained creativity label of the advertisements in \texttt{Creative-100}.

We utilize Amazon Mechanical Turk (MTurk) to collect the fine-grained annotations. For atypicality and originality, we follow \citet{modeling_determinants} and record responses about various statements (Table \ref{table:mturk_questions}). For creativity, we record a 5-scale score and convert it to a 3-scale one, aligning with other dimensions. Creativity is measured on a different scale because we believe there are subtle differences between ``very creative'' and ``somewhat creative''. To ensure the annotators actually understand the advertisement, we also include an annotation quality check question, asking annotators to choose the action after seeing a given advertisement (e.g., ``I should go to Chick-fil-A'' for advertisement A in Figure \ref{fig:intro}). In this question, five actions are given, with one correct action and four incorrect ones sampled from \texttt{Pitt-Ads}. Annotators get 96.88\% accuracy on this question, highlighting their accurate understanding of visual advertisements. Detailed annotation interface and instructions are in Figure \ref{fig:mturk}.

Unlike the traditional data annotation process that only collects a few annotations per data point, we choose to collect a larger number of annotations per data point. Thus, the predictive target is not a binary label (e.g., ``creative'' v.s. ``not creative'') but a distribution of human ratings (e.g., ``10\% chooses creative, 50\% neutral, 40\% not creative''). We believe this would better capture the inherent subjectivity of creativity judgment by showing the diverse perspectives from the annotators. In practice, we collect 25 annotations per advertisement image to approximate the true rating distribution within a certain error rate ~\cite{mchugh2012interrater}. 
Refer to Appendix \ref{sec:appendix_num_samples} for more details on how we choose the exact number of annotations; more dataset construction details are in Appendix \ref{sec:data_collection}.

\subsection{\texttt{Atypical-300}}
\label{sec:atypicality_data}
We also randomly sampled 300 advertisements (\texttt{Atypical-300}) from \texttt{Pitt-Ads} to specifically examine the atypicality dimension. In this subset, 185 instances (62\%) contain atypical object(s) and are assigned a positive label for “atypicality,”. The remaining 115 cases (38\%) are labeled negative.
\texttt{Atypical-300} is a larger and additional dataset similar to \texttt{Creative-100} but with only one dimension, atypicality. 
Different from \texttt{Creative-100}, each advertisement here only has three binary annotations on atypicality. Both ~\citet{modeling_determinants} and \texttt{Creative-100} (Appendix \ref{sec:appendix_connection}) show that atypicality has a positive correlation with creativity.  Thus, we include this dataset to further study VLM's ability to evaluate advertisement creativity.% across a wider range of ad images.

\input{sections/table_results}

%% file: sections/table_questions.tex
\begin{table}[t]
\footnotesize
\centering
    \resizebox{\linewidth}{!}{
    \begin{tabular}{p{1.4cm}|p{5.5cm}|p{1.8cm}}
        \toprule
        % \hline
        \textbf{Dimension} & \textbf{Questions} & \textbf{Answer} \\
         \midrule
         % \hline 			
        \multirow{3}{4em}{Atypicality} & The ad connected usually unrelated objects & \multirow{3}{6em}{agree (1), neutral (0), disagree (-1)} \\
        & The ad contained unusual connection &  \\
        & The ad brought unusual items together &  \\
        % \hline
        \midrule
        \multirow{3}{4em}{Originality} & The ad was out of the ordinary & \multirow{3}{6em}{agree (1), neutral (0), disagree (-1)} \\
        & The ad broke away from habit-bound and stereotypical thinking &  \\
        & The ad was unique &  \\
        \midrule
        Creativity & What is the overall level of creativity of this advertisement? 
        & integer (1-5) \\
        \bottomrule
        % \hline
    \end{tabular}
    }
    \caption{Questions in Amazon Mechanical Turk; full annotation interface and instructions is in Figure \ref{fig:mturk}.}
    \vspace{-6mm}
    \label{table:mturk_questions}
\end{table}

% \begin{table*}[t]
% \footnotesize
% \centering
%     \begin{tabular}{lll}
%         \toprule
%         % \hline
%         \textbf{Section} & \textbf{Questions} & \textbf{Answer Format} \\
%          \midrule
%          % \hline 			
%         Atypicality & \makecell[l]{The ad connected objects that are usually unrelated.\\The ad contained unusual connections.\\The ad brought unusual items together.}
%  & agree(1), neutral(0), disagree(-1)  \\
%         % \hline
%         \midrule
%         Originality & \makecell[l]{The ad was out of the ordinary.\\The ad broke away from habit-bound and stereotypical thinking.\\The ad was unique.} & agree(1), neutral(0), disagree(-1)  \\
%         \midrule
%         Creativity & \makecell[l]{What is the overall level of creativity of this advertisement?} 
%         & integer number (1-5) \\
%         \bottomrule
%         % \hline
%     \end{tabular}
%     \caption{Questions in Amazon Mechanical Turk}
%     \vspace{-6mm}
%     \label{table:mturk_questions}
% \end{table*}

%% file: sections/table_results.tex
\begin{table*}[t]
\small
\setlength{\tabcolsep}{3pt}
\centering
  \begin{tabular}{cc|cc|c|ccc}
    \toprule
    % \hline
    \multirow{3}{*}{\textbf{Dimension}} &
    \multirow{3}{*}{\textbf{Model }} &
      \multicolumn{2}{c}{\textbf{Distribution Modeling}} &
      \multicolumn{1}{c}{\textbf{Disagreement}} &
      \multicolumn{3}{c}{\textbf{Pairwise}} \\
        &&  {Rating Correlation} & {Distribution Divergence} & {} & {All} & {Easy} & {Hard} \\
        && {$\uparrow$ \textit{ R (p-value)}} & { $\downarrow$ \textit{KL Divergence}} & {$\uparrow$ \textit{R (p-value)}} & {$\uparrow$ \textit{F1}} & {$\uparrow$ \textit{F1}} & {$\uparrow$ \textit{F1}}  \\
        
    \midrule     
        \multirow{4}{*}{\shortstack{\textbf{Creativity} \\ (\texttt{Creative-100})}}
            & LLaVA-7B &  0.65 (.00$^*$) & 1.01 &  0.06 (.52) & 0.78 & 0.80 & 0.75 \\
            & LLaVA-13B   & 0.65 (.00$^*$) & \textbf{0.37} & \textit{nan} & 0.67 & 0.69	& 0.65\\
            & InternVL2-8B  & \textbf{0.72} (.00$^*$) & 1.45  &  \textbf{0.20} (.05) & 0.80 & 0.85 & 0.75 \\
            & GPT-4v   & - & - &  -0.04 (.72) & \textbf{0.97} & \textbf{0.98} & \textbf{0.96}\\
    \midrule     
        \multirow{4}{*}{\shortstack{\textbf{Originality} \\ (\texttt{Creative-100})}}
            & LLaVA-7B & 0.76 (.00$^*$) & 0.56 & 0.07 (.49) & 0.73 & 0.80 & 0.67 \\
            & LLaVA-13B   & 0.71 (.00$^*$)  & \textbf{0.30} & \textit{nan} & 0.67 & 0.68 & 0.67 \\
            & InternVL2-8B  & \textbf{0.82} (.00$^*$) & 0.57 & 0.11 (.27) & 0.69 & 0.71 & 0.68\\
            & GPT-4v   & - & - &  \textbf{0.15} (.13) & \textbf{0.87 } & \textbf{0.93} & \textbf{0.85}\\
    \midrule     
        \multirow{4}{*}{\shortstack{\textbf{Atypicality} \\ (\texttt{Creative-100})}}

            & LLaVA-7B &  0.75 (.00$^*$) & 0.40 & 0.17 (.08) & 0.76 & 0.81 & 0.72\\
            & LLaVA-13B   & 0.63 (.00$^*$) & \textbf{0.26} & \textit{nan} & 0.69 & 0.68 & 0.69\\
            & InternVL2-8B  & \textbf{0.78} (.00$^*$) & 0.40 & \textbf{0.24} (.02$^*$) & 0.78 & 0.81 & 0.75 \\
            & GPT-4v   & - & - &  -0.07 (.47) & \textbf{0.89 } & \textbf{0.94} & \textbf{0.80}\\
            
    \midrule  
    \midrule  
        \multirow{4}{*}{\shortstack{\textbf{Atypicality} \\ (\texttt{Atypical-300})}}
            & LLaVA-7B &  0.21	(.00$^*$) & 0.32 &  0.01 (.92) &  0.79  & - & - \\
            & LLaVA-13B  & 0.17 (.00$^*$) & \textbf{0.26} & -0.05 (.43) & 0.66 & - & - \\
            & InternVL2-8B  & \textbf{0.23} (.00$^*$) & 0.35 &  \textbf{0.02} (.75) & 0.77 & - & - \\
            & GPT-4v   & - & - &  -0.00 (.96) & \textbf{0.90}  & - & - \\

    % \hline
    \toprule
  \end{tabular}
  
  \caption{\textbf{Bold}: best-performing models; $^*$: statically significant results ($\alpha = 0.05$); \textit{nan}: disagreement predictions are uniform, making correlation test fail; ``-'' in GPT-4v rows: no distribution modeling task is done due to budget constraint; ``-'' in Pairwise columns: the classification of easy and hard is not available in \texttt{Atypical-300}.
  % \textbf{We can see promising performance in both Distribution Modeling and Pairwise Task, with Disagreement Prediction remains challenging.}
  }
  \label{table:mturk_results}
  \vspace{-5mm}
\end{table*}

%% file: sections/4_exp.tex
\section{Experimental Setup}
\subsection{Models}
We experiment with open-sourced VLMs, InternVL2-8B \cite{chen2024fargpt4vclosinggap}, LLaVA-7B and 13B \cite{li2024llavanext-strong}, and close-sourced VLMs, GPT4-v \cite{openai2024gpt4}. All experiments are done with zero-shot prompting with inference framework vllm \footnote{\href{https://docs.vllm.ai/en/v0.6.1.post1/getting_started/installation.html}{vllm 0.6.1.post1}} \cite{kwon2023efficient} . More details are in Appendix \ref{sec:appendix_experiment_details}.

\subsection{Task Formulation}
We define the following three tasks for each dimension (i.e., creativity, originality, and atypicality): 
\label{sec:distribution_task}
\textbf{Distribution Modeling} is designed to evaluate the model's ability to simulate human group behavior when it comes to creativity ratings. In practice, we prompt VLMs multiple times with high temperatures to get the same number of VLM outputs as the number of annotators (more details in Appendix \ref{sec:multi_run}). In this way, we simulate a ``group behavior'' instead of a single-point judgment about the level of creativity in the advertisement. To evaluate the quality of this simulation, we use two metrics: Spearman's correlation between the average rating from humans and that from VLMs, and the average KL Divergence between the human rating distribution and that from VLMs. These two results are in \textit{Rating Correlation} and \textit{Distribution Divergence} columns in Table \ref{table:mturk_results}. 

% \paragraph{Disagreement Prediction}
\label{sec:disagreement_task}
\textbf{Disagreement Prediction} tries to capture the annotator’s level of disagreement, which is important in domains like marketing. An advertisement with high creativity ratings and minimal disagreement is desired and could have a more positive impact on the product. In practice, we directly prompt VLMs to predict the level of disagreement (low, middle, or high) for each scoring dimension. We then compute Spearman's correlation between the prediction and the standard deviation of human ratings. This metric studies the level of creativity ambiguity of the advertisements. A very creative advertisement will have a low disagreement rate with a high creativity score. The results are in \textit{Disagreement} column in Table \ref{table:mturk_results}.

% \paragraph{Pairwise Preference}
\textbf{Pairwise Preference} aims at evaluating the model's ability to correctly pick the more creative advertisement out of two advertisements, given that an absolute rating of creativity can be hard when there is no reference.
For each scoring dimension, we include all advertisement pairs with average human ratings differences greater than 0.5. For \texttt{Creative-100}, we have 938, 2708, and 2631 pairs in creativity, originality, and atypicality; for \texttt{Atypical-300}, we sampled 1000 image pairs from 300 images due to constraints in computation resources. The results are evaluated by F1 score and are shown in \textit{Pairwise} column in Table \ref{table:mturk_results}.

%% file: sections/5_results.tex
\section{Results}
\textbf{Promising Results in Distribution Modeling}
For all dimensions in \texttt{Creative-100}, the correlations between average human and VLM ratings are both high and statistically significant, with InternVL2-8B being the best-performing model in every dimension. However, the correlations are much lower in \texttt{Atypical-300}. We hypothesize this is due to the small annotation size (3 per advertisement) in \texttt{Atypical-300}, which can easily be biased by one annotation data point, leading to an overall more noisy distribution. Distribution Divergence also shows promising results, with the lowest divergence achieved by the LLaVA-13B model. Cross-dataset disparity is also much lower, where the KL divergence is similar for the atypicality in both datasets. 

It is also worth noting that different distributional metrics result in different best-performing models, with InternVL2-8B tops Rating Correlation and LLaVA-13B wins KL Divergence. Further analysis reveals that LLaVA-13B tends to produce overall lower scores (average 2.19, out of 3) for all three dimensions compared to InternVL2-8B (average 2.31, out of 3). The MTurk human annotators produced a similarly lower score of 1.94 out of 3 on average. We note that KL Divergence focuses more on the absolute difference between two distributions of scores (LLaVA-13B scores are similarly low compared to humans), while the Rating Correlation cares more about the relative ranking of scores. This indicates that these two metrics measure different aspects of the score difference between human and model scorers.
Output examples and reasoning text word cloud in Appendix \ref{sec:appendix_distribution_output}.

\textbf{Disagreement Prediction Remains Challenging} For all scenarios in Disagreement Prediction (except for Atypicality in \texttt{Creative-100}, with InternVL2-8B), disagreement predictions have no statistical correlation with human rating standard deviations. For LLaVA-13B, all outputs are ``middle'', making the correlation result \textit{nan}. Both of these findings suggest that using VLM as a group-opinion synthesizer remains challenging\footnote{We also calculated the correlation between standard deviations of model predictions and that of human ratings, but the correlations are all near zero.}. We believe the main reason for such poor performance is the complexity involved in understanding and predicting disagreement. Understanding disagreement requires not only correctly interpreting the advertisement content, but also understanding how different humans would decode the meaning from different perspectives.
Since many advertisements require complex background knowledge (e.g., some require rarely known background knowledge, which might make people disagree highly on whether an advertisement is creative), correctly understanding such necessity is challenging but critical in correctly predicting the disagreement. This process involves multi-hop (or even hierarchical) reasoning, and it remains challenging for VLMs. We hypothesize that adding the related background knowledge of the advertisements, or decomposing the multi-step reasoning needed for the disagreement task, might improve model performance, but we leave this for future work to explore.
\textbf{Great Performance in Pairwise Preference} Results from the pairwise preference task are very impressive, with the best-performing GPT-4v achieving more than 0.9 F1 score. We also further analyze the performance by dividing image pairs into ``easy'' and ``hard'' pairs. If the difference between average human ratings of an image pair is higher than the median difference of all image pairs, it is considered an ``easy'' pair. In other words, those two images have a relatively higher difference in one of the dimensions. 
As shown in Table \ref{table:mturk_results}, all VLMs perform better in easy tasks than hard ones.

\textbf{Smaller Models' Superior Performance} LLaVA-7B and InternVL2-8B consistently outperform LLaVA-13B in Rating Correlation and Pairwise Preference, both of which are ranking-based evaluations. Based on the error analysis (Appendix \ref{sec:appendix_error_analysis}), we believe this can be explained by ranking tasks requiring higher reasoning capability from the language part of the VLM. The language parts of those two smaller models are Mistral-7B and InternLM2.5-7B-Chat, both having a higher ranking on the HuggingFace Open LLM LeaderBoard\footnote{\href{https://huggingface.co/spaces/open-llm-leaderboard/open_llm_leaderboard}{HuggingFace LLM Leaderboard}} in reasoning tasks compared to the language part (Vicuna-13B) of LLaVA-13B.

%% file: sections/6_conclusion.tex
\section{Conclusion}
%\minisection{Conclusion} 
We present a case study of using SoTA VLMs to evaluate creativity in advertisements. Inspired by marketing research, we collect fine-grained human annotations for atypicality, originality, and overall creativity, with enough annotations for every data point to capture the subjective nature of creativity evaluation. We also propose several new tasks, including distribution modeling and disagreement prediction, which specifically test VLMs' ability to deal with subjectivity. We find that SoTA VLMs achieve promising results in the pairwise comparison task while still struggling with disagreement prediction and distribution modeling. Our work opens the opportunity for the automatic assessment of advertisement creativity by providing a benchmark and metrics.

Looking ahead, although GPT-4v achieves impressive results in the pairwise comparison task, we still see smaller, open-source models like LLaVA-7B underperforming. Since smaller models are more commonly used in downstream applications, future work could focus on improving their performance.

\section{Limitations}
One obvious limitation is the size of our dataset. The fine-grained creativity annotation only consists of 100 advertisement images. Two bottlenecks that lead to such a limited number are the budget and annotation quality. Since we want to explore distribution modeling, we need more annotations than typical machine learning tasks, leading to a huge budget requirement. Also, because of the dataset size, we are not able to conduct fine-tuning experiments, as we only have 100 images with high-quality annotations. We will leave exploration of fine-tuning experiments to future work.

Another limitation of the paper is the design of simulating ``group behavior'' by prompting a VLM 25 times with the same prompt, we recognize the simplicity of how we prompted the VLMs to make predictions. However, creating 25 different prompts for each advertisement could complicate the analysis and results. For example, certain prompts could disproportionately increase the likelihood of a ``creative'' label compared to others. One interesting approach to generating prompts more meaningfully could involve exploring persona prompts to simulate multiple annotators' behavior. However, we believe that starting with a straightforward approach using the sampling strategy is essential. Therefore, we leave the use of 25 distinct persona prompts for future work. 

Finally, due to hardware constraints, we only experiment with VLMs in the 7B to 13B range when much larger models, such as LLaVA-34B, are available. We will leave more extensive prompt tuning and model selections to future work.

\section*{Ethical Consideration}
As a majority of our annotators are located in the U.S, there are natural biases in our annotation. We have plans to expand the annotation to other platforms (e.g., LabInTheWild \citet{LabintheWild}) where a more diverse set of annotators is available. Future work could also explore alternative prompting approaches to simulate group behavior or conduct a demographic analysis of human annotations, which could check whether VLM holds opinions that are comparable to those of particular groups.

\section*{Acknowledgment}
We appreciate the valuable feedback and constructive suggestions provided by the anonymous reviewers and researchers at the Pitt NLP group. 
This research was supported in part by the University of Pittsburgh Center for Research Computing and Data, RRID:SCR\_022735, through the resources provided. Specifically, this work used the H2P cluster, which is supported by NSF award number OAC-2117681.

%% file: sections/appendix.tex
\appendix

\section{Ads Dataset}
\label{sec:appendix_dataset}

\subsection{Terminology}
\label{sec:appendix_terminologies}
In the original study \cite{modeling_determinants} that proposed the breakdown of creativity, they use the term ``synthesis'' instead of atypicality. However, they defined ``synthesis'' as ``\textit{...combine, connect, or blend normally unrelated objects or ideas}'' which is very similiar to ``atypicality''. We use the term ``atypicality'' because that is the term used in the \texttt{Pitt-Ads} Dataset and want to keep the terminology aligned. %From now on, we refer to both of them as ``atypicality''.

\begin{figure}
    \centering
        \includegraphics[width=0.4\textwidth]{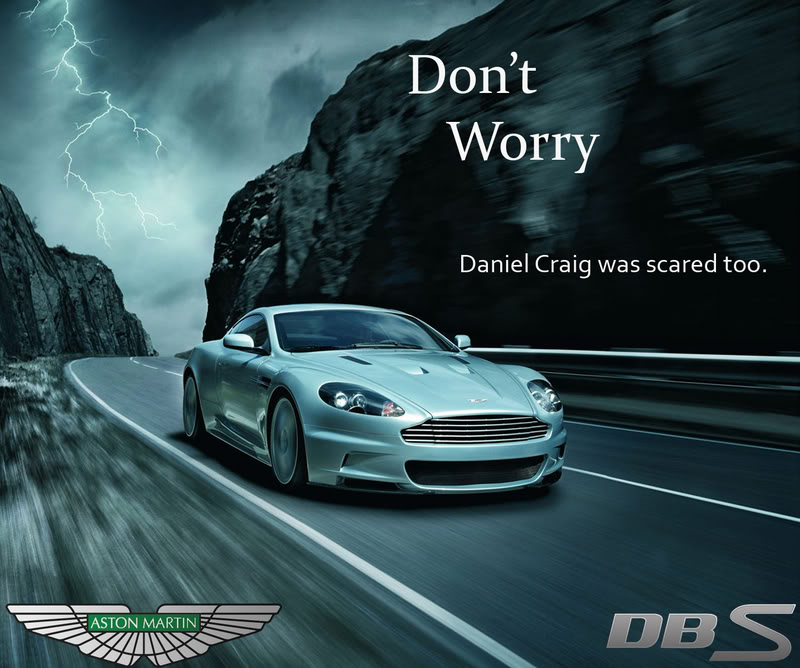}
    \caption{Example of original but not so atypical advertisement.}
    \label{fig:ad_example}
    % \vspace{-8mm}
\end{figure}

In terms of their differences, Atypicality is intrinsic to the ad itself, e.g., it contains physically impossible scenes such as people dancing under the ocean; originality is extrinsic and is the result of comparing with other existing ads in the same domain, e.g., a car ad without any image of cars on it is considered original. These two aspects are complementary: some ads are original but not atypical, e.g., a regular car and a typical scenario, but original text that involves a reference to pop culture. The advertisement in Figure \ref{fig:ad_example}has an average of 1.56 in atypicality and 2.0 in originality (scale 1 to 3).

\begin{figure}
    \centering
        \includegraphics[width=0.4\textwidth]{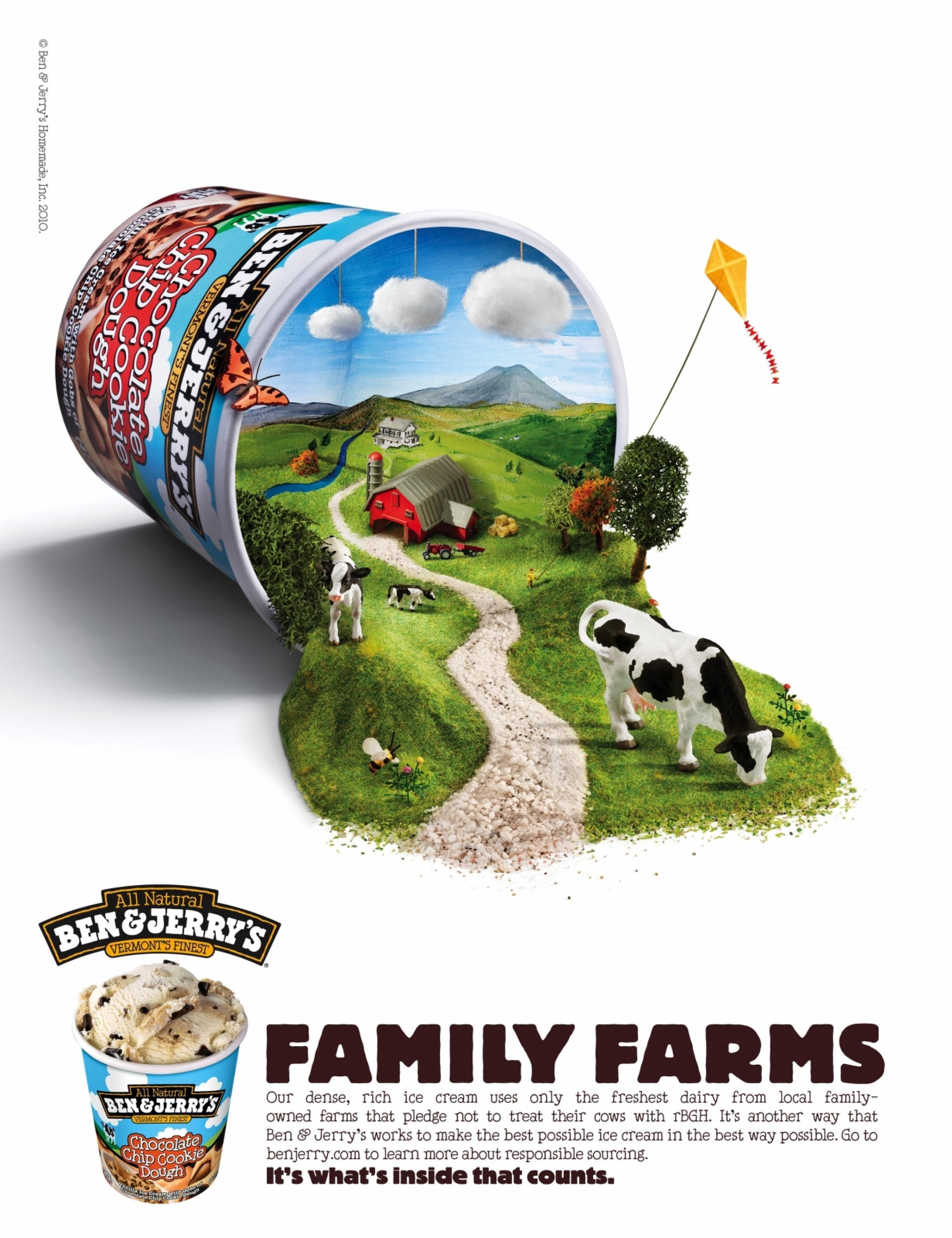}
    \caption{Example of atypical advertisement.}
    \label{fig:ad_example_atypicality}
    % \vspace{-8mm}
\end{figure}

\subsection{Atypicality and Originality as Selected Dimensions of Creativity}
\label{sec:app-selected-dimensions}
The original marketing research \cite{modeling_determinants} conducted an exploratory factor analysis (EFA) on five dimensions of divergence in ads creativity (numbers in parenthesis are loading factors of EFA, higher means more correlated with creativity): originality (0.23), synthesis (0.23), artistic value (0.20), flexibility (0.18), and elaboration (0.15). As mentioned previously, their definition of synthesis is very similar to ``atypicality'' as defined in \texttt{Pitt-Ads} Dataset. Therefore, we keep the highest-scoring two aspects (originality and synthesis/atypicality) in our study. 

\subsection{Sampling Process}
% \label{sec:appendix_dataset}
The original \texttt{Pitt Ads} dataset consists of 38 topics in total \cite{Hussain2017AutomaticUO}. We group those topics into 10 ``topic groups'': clothing, food, automobile, beauty, leisure, electronics, drinks, service, non-commercial, and healthcare. From each topic group, we sampled 10 ads, which makes up the 100 images in \texttt{Creativity-100}.

\paragraph{Details about ``topic groups''}
\begin{itemize}
    \item food: ``Restaurants, cafe, fast food'', ``Chocolate, cookies, candy, ice cream'', ``Chips, snacks, nuts, fruit, gum, cereal, yogurt, soups'', ``Seasoning, condiments, ketchup''
    \item pet: ``Pet food''
    \item drinks: ``Alcohol"'', ``Coffee, tea"'', ``Soda, juice, milk, energy drinks, water"''
    \item automobile: ``Cars, automobiles'' 
    \item electronics: ``Electronics'' 
    \item service: ``Phone, TV and internet service providers'', ``Financial services , ``Security and safety services'' , ``Software'' , ``Other services'' 
    \item education: ``Education''
    \item beauty: ``Beauty products and cosmetics''
    \item healthcare: ``Healthcare and medications''
    \item clothing: ``Clothing and accessories''
    \item home: ``Baby products'' , ``Cleaning products'' , ``Home improvements and repairs'' , ``Home appliances'' ]
    \item leisure: ``Games and toys'' , ``Vacation and travel'' , ``Media and arts'' , ``Sports equipment and activities'', ``Gambling'' 
    \item shopping: ``Shopping''
    \item non-commercial: ``Environment, nature, pollution, wildlife'', ``Animal rights, animal abuse'', ``Human rights'', ``Safety, safe driving, fire safety'', ``Smoking, alcohol abuse'', ``Domestic violence'', ``Self esteem, bullying, cyber bullying'', ``Political candidates'', ``Charities''
\end{itemize}

\begin{figure*}
    \centering
        \includegraphics[width=\textwidth]{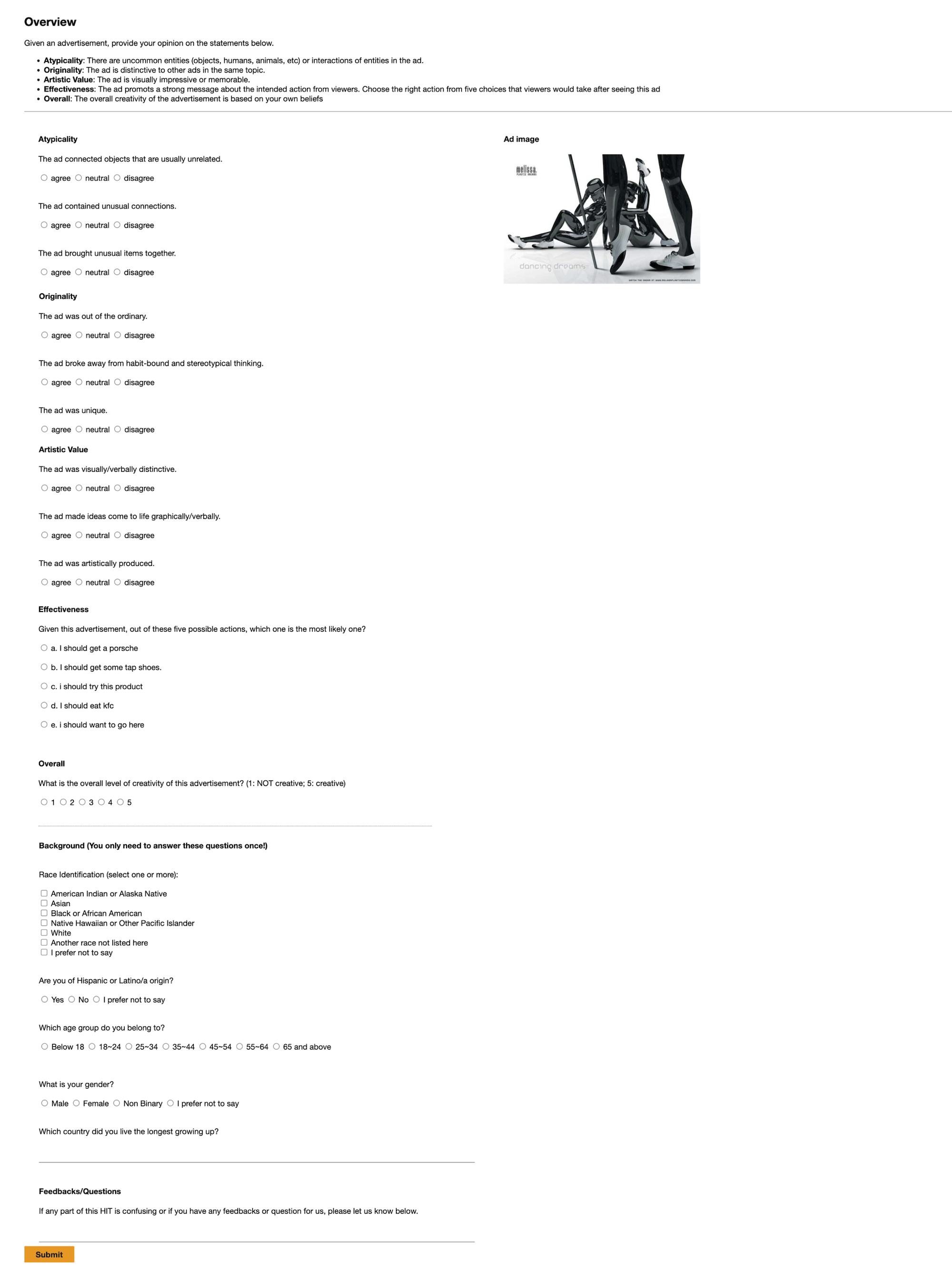}
    % \caption{Task formulation}
    \caption{Amazon Mechanical Turk interface.}
    \label{fig:mturk}
    % \vspace{-5mm}
\end{figure*}

\section{Data Collection and Post Processing}
\label{sec:data_collection}
\subsection{Amazon Mechanical Turk Details}
% \label{sec:}
\label{sec:appendix_mturk}
\paragraph{Payment for worker}
Each HIT receives \$0.5 compensation (estimated \$15/hour).

\paragraph{Annotation interface}
See Figure \ref{fig:mturk} for the annotation interface. Note that there is a section ``artistic values''. We dropped that section in the later parts of the experiment because 1) it is very subjective and could be further broken down into more fine-grained subcategories, and 2) to keep our focus on atypicality and originality.

\begin{figure}
    \centering
        \includegraphics[width=0.4\textwidth]{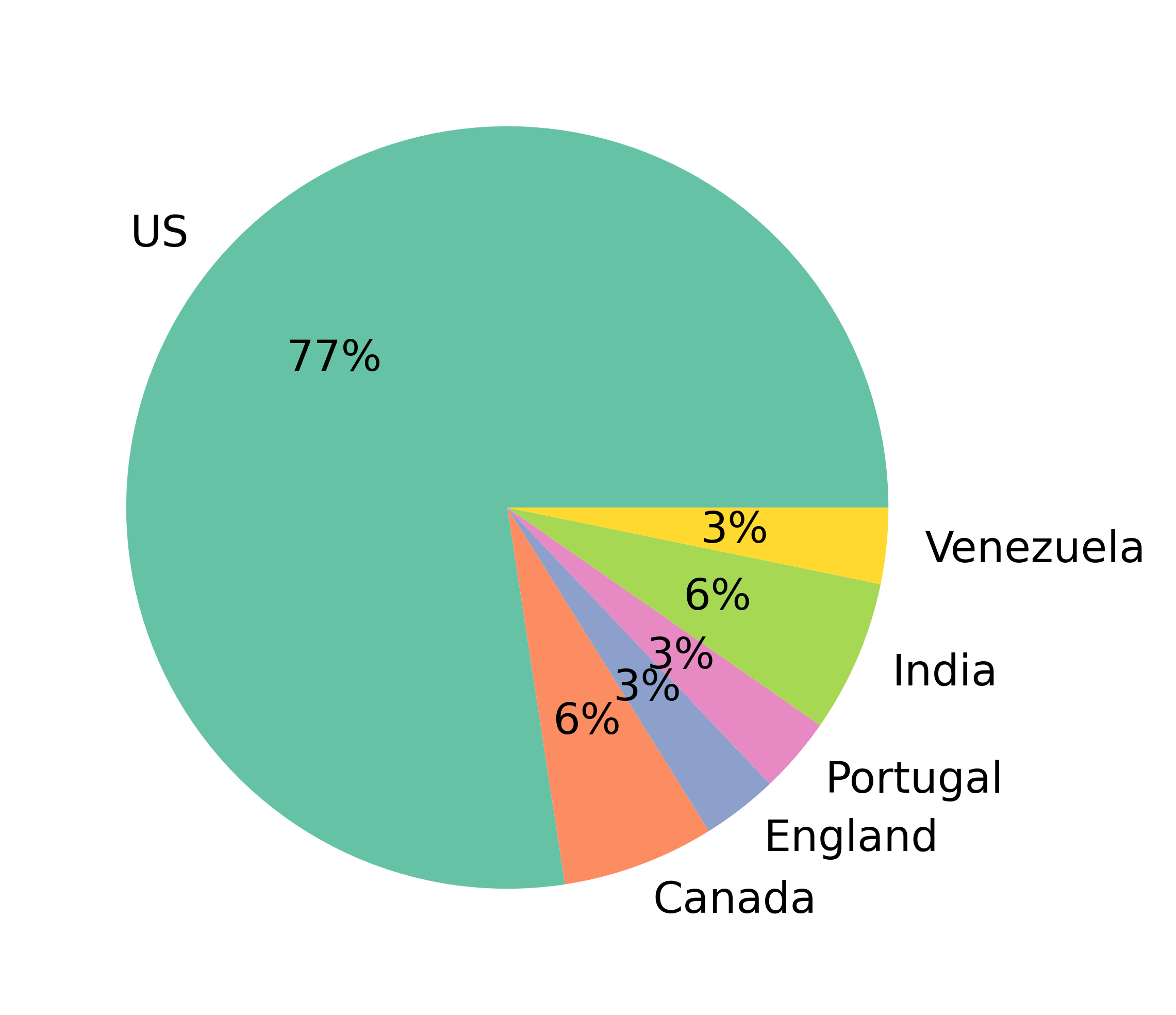}
    \caption{Distribution of workers' response to ``In which country did you live the longest time so far?''}
    \label{fig:mturk_countries}
    % \vspace{-8mm}
\end{figure}

\begin{figure}
    \centering
        \includegraphics[width=0.4\textwidth]{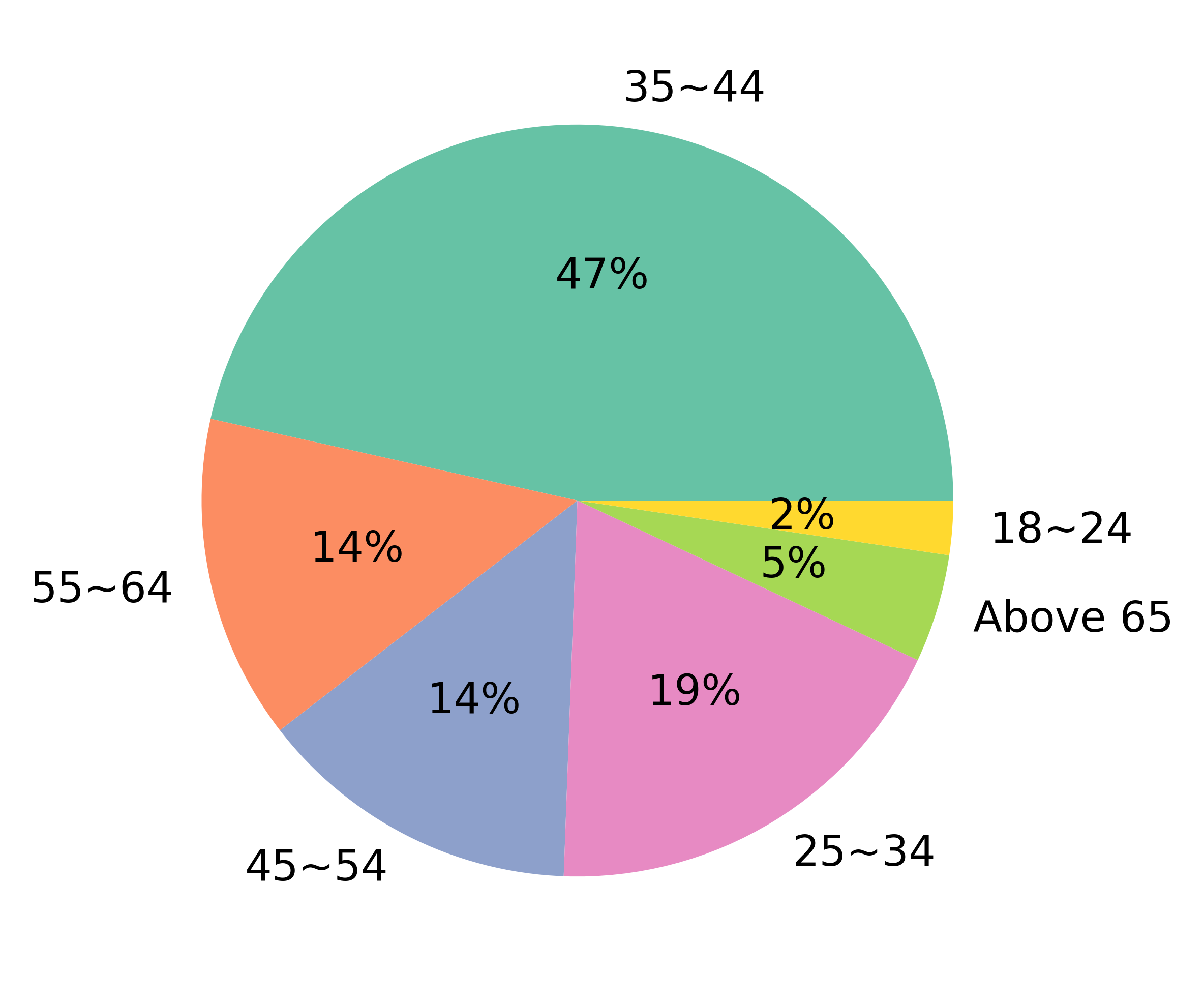}
    \caption{Distribution of workers' response to ``What is your age?''}
    \label{fig:mturk_age}
    % \vspace{-8mm}
\end{figure}

\begin{figure}
    \centering
        \includegraphics[width=0.4\textwidth]{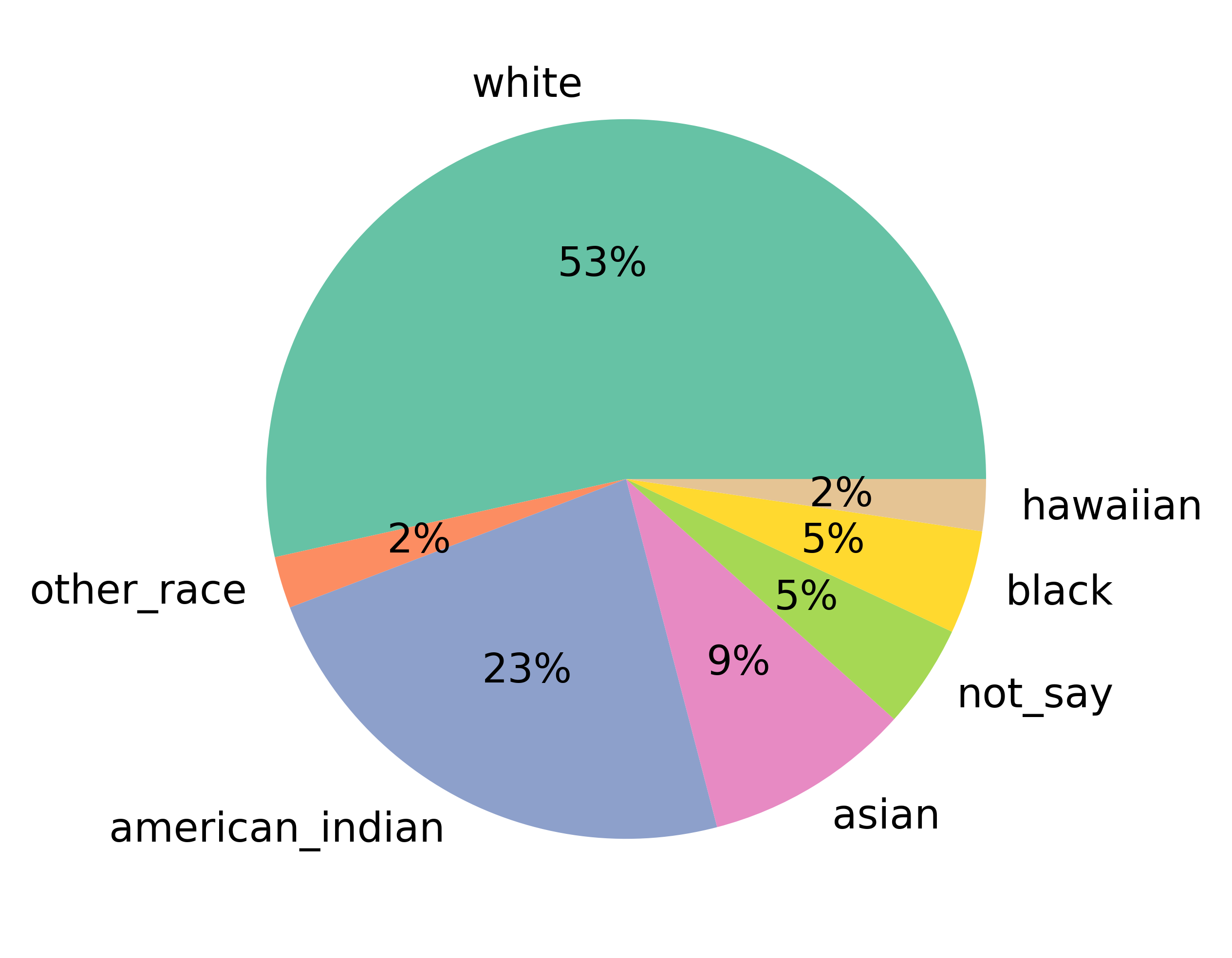}
    \caption{Distribution of workers' response to ``Race Identification (select one or more)''}
    \label{fig:mturk_race}
    % \vspace{-8mm}
\end{figure}

\begin{figure}
    \centering
        \includegraphics[width=0.4\textwidth]{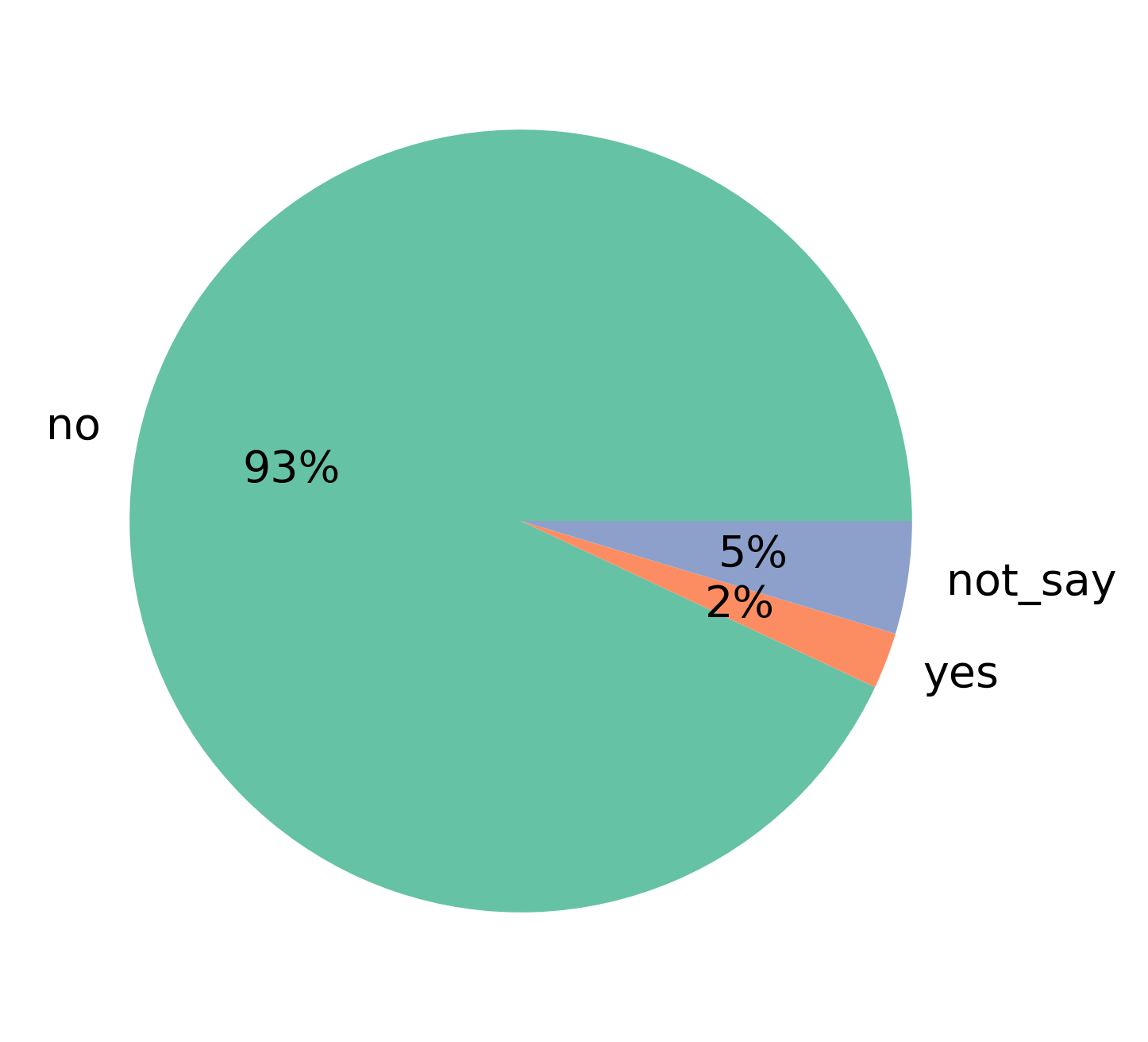}
    \caption{Distribution of workers' response to ``Are you of Hispanic or Latino/a origin?''}
    \label{fig:mturk_ethnicity}
    % \vspace{-8mm}
\end{figure}

\begin{figure}
    \centering
        \includegraphics[width=0.4\textwidth]{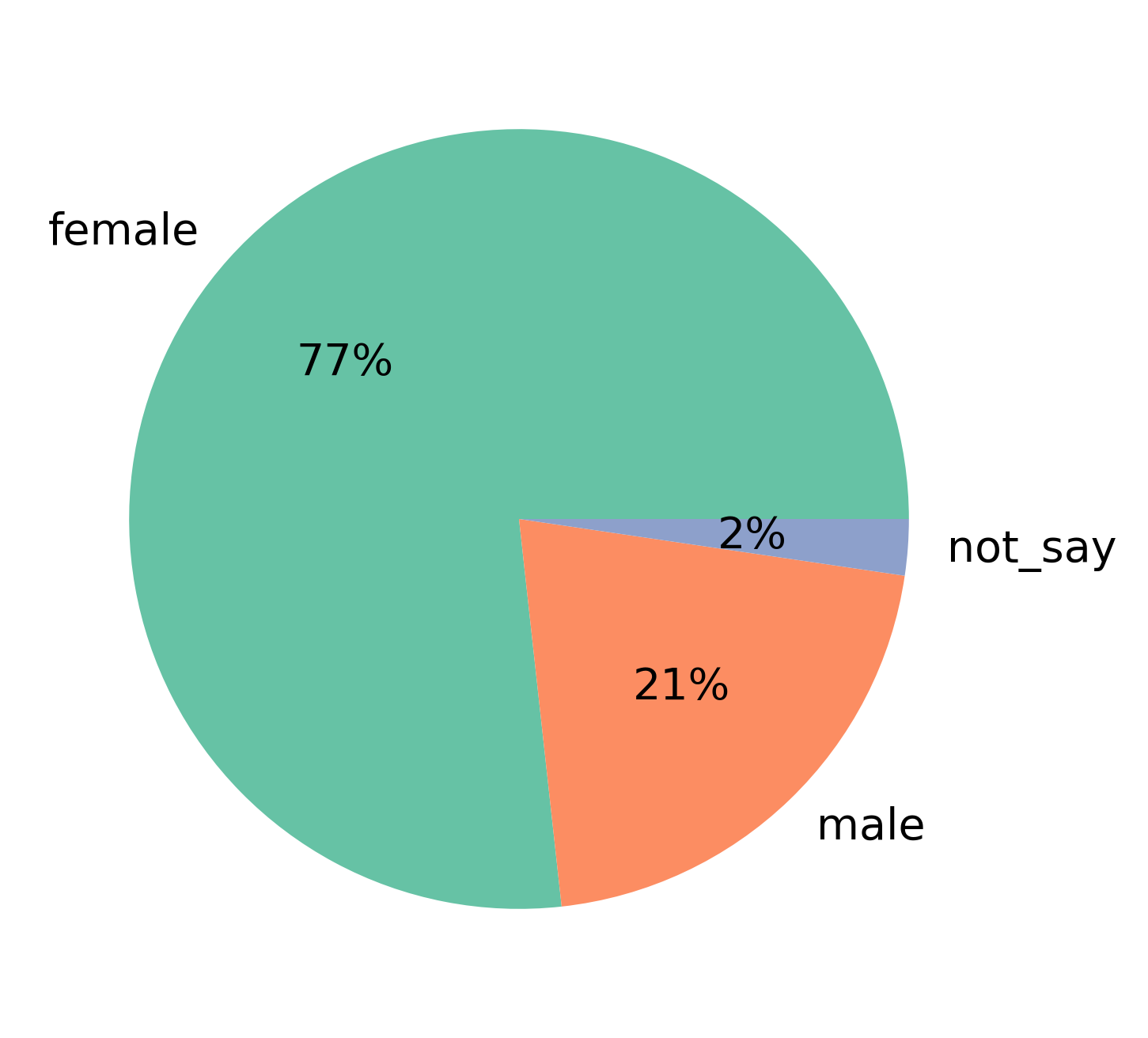}
    \caption{Distribution of workers' response to ``What is your gender?''}
    \label{fig:mturk_gender}
    % \vspace{-8mm}
\end{figure}

In total, 43 workers contributed to our task and finished 2500 HITs. Their background can be found in Figure \ref{fig:mturk_countries}, \ref{fig:mturk_age}, 
\ref{fig:mturk_race}, \ref{fig:mturk_ethnicity}, and \ref{fig:mturk_gender}. As we can see, the annotators are strongly skewed towards the US-based, white, female, middle age group, which should be kept in mind when applying our methodology when it comes to people from another background. 

\subsection{Number of Samples for Distribution Task}
\label{sec:appendix_num_samples}
\begin{figure}
    \centering
        \includegraphics[width=0.4\textwidth]{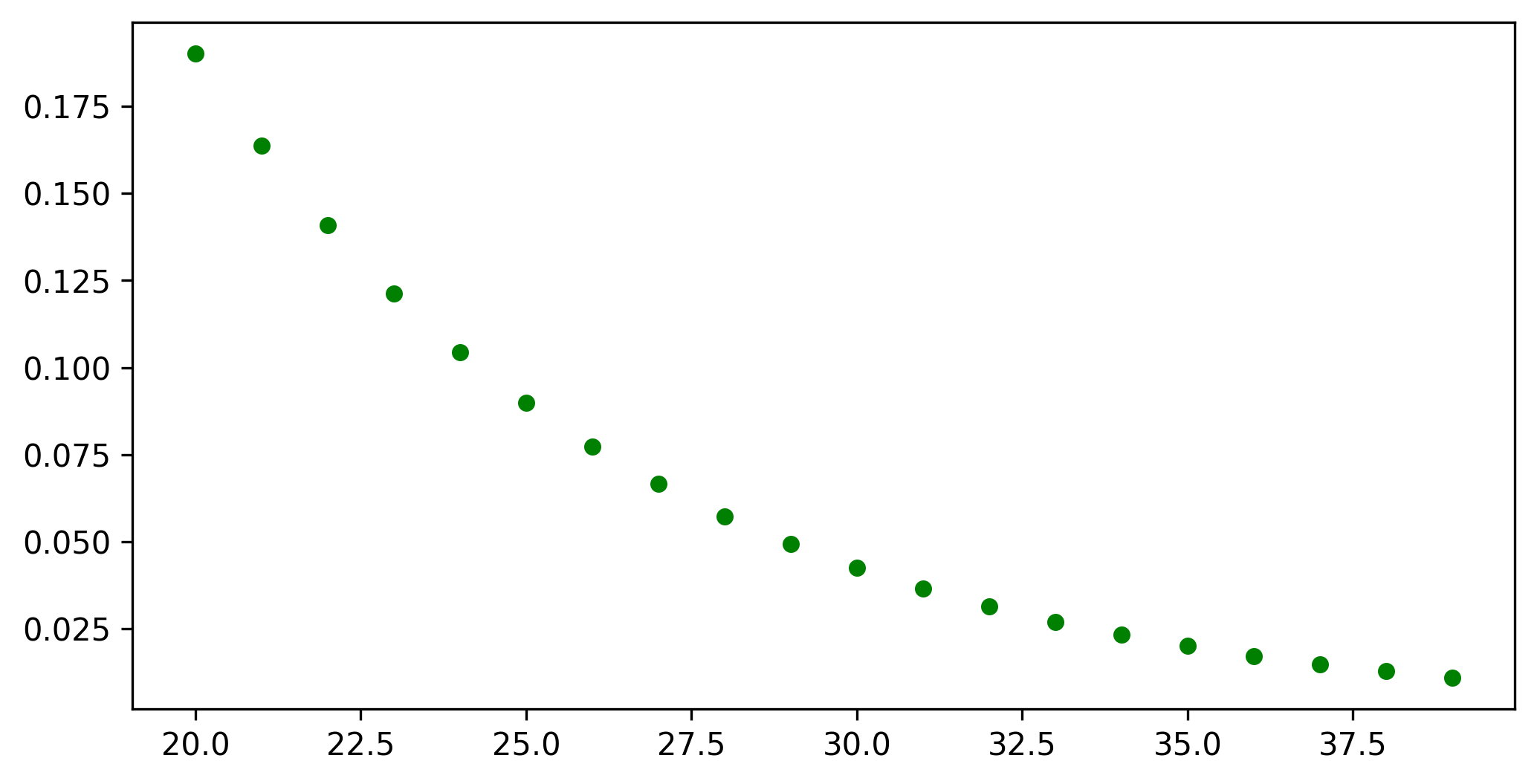}
    \caption{Upper-bound of the error based on calculation.}
    \label{fig:sample_size_error}
    % \vspace{-8mm}
\end{figure}

Following previous works \cite{mchugh2012interrater, cheng2024every}, the number of samples required to approximate the real distribution can be calculated as follows: 
\[ P(D_{KL}(g_{n, k}||f) > \epsilon) \leq e^{-n\epsilon} \left[  \frac{3c_1}{c_2}\sum_{i=0}^{k-2} {k_{i-1}} (\frac{e\sqrt{n}}{2\pi})^i \right] \]
$c_1$ and $c_2$ are constant values (based on \cite{mchugh2012interrater} $c_1 = 2, c_2 = \frac{\pi}{2}$), k is the number of categories in the categorical distribution (in our case, $k = 3$), and n is the number of samples. If we fix the left-hand side to be less than 0.1, we would get $n$ has to be 25 (see Figure \ref{fig:sample_size_error}).

\subsection{Label Processing}
\label{sec:appendix_label_proc}

We process the annotation by first converting the categorical data to numerical values. For atypicality and originality, we code agree, neutral, and disagreement choices as 1, 0, and -1. As there are three subquestions for both atypicality and originality, we simply add up the three scores from each dimension and get one accumulated score for each. For overall creativity, we keep the raw score (an integer number between 1 and 5). Thus each annotation data point consists of three integer scores, corresponding to atypicality, originality, and overall creativity.

We then normalize the score by individual annotators to mitigate the differences in people's rating preferences. In particular, for each score dimension, we group the scores provided by each annotator and standardize them (subtract mean and divide by standard deviation). We then map the standardized score to an integer (1, 2, or 3) by dividing the standardized score interval into three bins.

\subsection{Connection between atypicality and creativity}
\label{sec:appendix_connection}
After analyzing the fine-grained creativity data we collected (Sec.~\ref{sec:fine_grained_creativity_data}), we find out that the Pearson R correlation between the normalized atypicality and overall creativity score is 0.4017 ($p<0.01$), a positive correlation (the sample size is 2500: 100 ads with 25 annotations each). Therefore, it makes sense to evaluate the same methodology on data with only atypicality annotation to prove its effectiveness at a larger scale. 

\section{Experiment Details}
\label{sec:appendix_experiment_details}
\subsection{Configurations}
\begin{itemize}
    \item Temperature: 0.75 (for distribution prediction) and 0.01 (for disagreement prediction)
    \item Max New Token: 256
    \item Model Checkpoint
    \begin{itemize}
        \item GPT-4: \texttt{gpt-4-vision-preview}
        \item LLaVa-7B: \texttt{llava-v1.6-mistral-7b-hf}
        \item LLaVa-13B: \texttt{llava-v1.6-vicuna-13b-hf}
        \item InternVL2-8B: \texttt{InternVL2-8B-MPO}
    \end{itemize}
    \item Number of pairwise samples (\% of label ``1'')
    \begin{itemize}
        \item creativity: 938 (46\%)
        \item atypicality: 2631 (51\%)
        \item originality: 2708 (50\%)
    \end{itemize}
    \item Device: single card NVIDIA L40S GPU
\end{itemize}

\subsection{Running Time}(Roughly, all tasks combined) 
\begin{itemize}
    \item \texttt{Creative-100}
    \begin{itemize}
        \item GPT4-v: 1 hr
        \item LLaVA-7B: 4 hr
        \item LLaVA-13B: 5 hr
        \item InternVL: 4.5 hr
    \end{itemize}
    \item \texttt{Atypical-300} (atypical data only)
    \begin{itemize}
        \item GPT4-v: 0.2 hr
        \item LLaVA7B: 0.5 hr
        \item LLaVA13B: 0.5 hr
        \item InternVL: 0.5 hr
    \end{itemize}
\end{itemize}

\subsection{Multiple Run Results}
\label{sec:multi_run}
For the distribution modeling task, since there is inherent randomness in the modeling process (we set the temperature for LLM generation to 0.75), we run the same setting for 3 times to see if the randomness will have a huge impact. As shown in Table \ref{table:multi-run results}, for the same task, same model, the performance is still within reasonable margin. We believe it is safe to believe that the results is not affected by randomness in LLM.

\input{sections/table_multi_run}

\subsection{VLM Prompts - Creativity}
\paragraph{Distribution Modeling} 

\textit{How creative is this advertisement? Give your answer that follows these guidelines:\\
- Choose your answer from [not creative, neutral, very creative], answer 1 for not creative, 2 for neutral, and 3 for very creative \\
- Show your reasons and give a final answer (in a single number), in the following format: ``reasoning:\{reasoning\}, answer: \{score\}''}

\paragraph{Disagreement}
\textit{You are a marketing analyst tasked with predicting the audience's perception of an advertisement's creativity. In next week, 25 individuals will rate this ad's creativity on a scale of 1 to 3. Your goal is to predict the level of disagreement among these ratings. As a reference, ads that are more challenging to understand generally result in higher disagreement, while simpler ads lead to greater consensus.} \\
\textit{Provide your assessment using a single score:\\
1: Low disagreement (high agreement),\\
2: Neutral disagreement,\\
3: High disagreement (low agreement).\\
Your response should follow this format:\\
``answer: \{score\}; explanation: \{reasoning\}''}

\paragraph{Pairwise}
\textit{Here are two images of advertisement. Which one is more creative? Answer 1 for the one on the left and 2 for the one on the right. Give your answer in the following format: ``explanation: \{reasoning\}; answer: \{choice\}''}

\subsection{VLM Prompts - Atypicality}
\paragraph{Distribution Modeling}
\textit{How atypical is this advertisement? Give your answer that follows these guidelines:\\
- Atypical advertisement either connected objects that are usually unrelated, contained unusual connections, or brought unusual items together;\\
- Choose your answer from [not atypical, neutral, very atypical], answer 1 for not atypical, 2 for neutral, and 3 for very atypical\\
- Show your reasons and give a final answer (in a single number), in the following format: ``reasoning: \{reasoning\}, answer: \{score\}; ''}

\paragraph{Disagreement} 
\textit{You are a marketing expert task with predicting how audiences perceive the atypicality of an advertisement. Atypicality refers to ads that connect objects typically unrelated, feature unusual connections, or bring unexpected items together. In next week, 25 individuals will rate this ad's atypicality on a scale of 1 to 3. Your task is to predict the level of disagreement among their ratings. As a reference, Ads that are more difficult to interpret tend to generate higher disagreement, whereas more straightforward ads lead to greater agreement.\\
Provide your answer using a single score:\\
1: Low disagreement (high agreement),\\
2: Neutral disagreement,\\
3: High disagreement (low agreement).\\
Format your response as follows:\\
``answer: \{score\}; explanation: \{reasoning\}''}

\paragraph{Pairwise}
\textit{Here are two images of advertisement. Which one is more abnormal and unusual? Answer 1 for the one on the left and 2 for the one on the right. Give your answer in the following format: ``explanation: \{reasoning\}; answer: \{choice\}''}

\subsection{VLM Prompts - Originality}
\paragraph{Distribution Modeling}
\textit{How novel is this advertisement? Give your answer that follows these guidelines:\\
- Novel advertisement either is out of the ordinary, break away from habit-bound and stereotypical thinking, or is unique;\\
- Choose your answer from [not novel, neutral, very novel], answer 1 for not novel, 2 for neutral, and 3 for very novel\\
- Show your reasons and give a final answer (in a single number), in the following format: ``reasoning: \{reasoning\}, answer: \{score\};''}

\paragraph{Disagreement} 
\textit{You are a marketing analyst tasked with predicting how audiences perceive the novelty of an advertisement. Novelty refers to ads that are out of the ordinary, break free from stereotypical thinking, or exhibit uniqueness. In next week, 25 individuals will rate this ad's novelty on a scale of 1 to 3. Your goal is to predict the level of disagreement among their ratings. Ads that are harder to interpret typically result in higher disagreement, while clearer ads lead to more agreement.\\
Provide your assessment using a single score:\\
1: Low disagreement (high agreement),\\
2: Neutral disagreement,\\
3: High disagreement (low agreement).\\
Format your response as follows:\\
``answer: \{score\}; explanation: \{reasoning\}''}

\paragraph{Pairwise}
\textit{Here are two images of advertisement. Which one is more unique compared with other ads in the same product category? Answer 1 for the one on the left and 2 for the one on the right. Give your answer in the following format: : ``explanation: \{reasoning\}; answer: \{choice\}''}

\subsection{\texttt{Atypical-300} Prompts (atypicality only)}
\paragraph{Distribution Modeling}
\textit{How atypical is this advertisement? Give your answer that follows these guidelines:\\
- Atypical advertisement either connected objects that are usually unrelated, contained unusual connections, or brought unusual items together;\\
- Choose your answer from [not atypical, neutral, very atypical], answer 0 for not atypical and 1 for very atypical\\
- Show your reasons and give a final answer (in a single number), in the following format: ``reasoning: \{reasoning\}, answer: \{score\};''}

\paragraph{Disagreement} 
\textit{You are a marketing expert analyzing how audiences perceive the atypicality of an advertisement. Atypicality refers to ads that connect objects typically unrelated, feature unusual connections, or bring unexpected items together. Imagine 25 individuals have rated the ad's atypicality either 0 or 1. Your task is to determine the level of disagreement among their ratings. Ads that are more difficult to interpret tend to generate higher disagreement, whereas more straightforward ads lead to greater agreement.\\
Provide your answer using a single score:\\
0: Low disagreement (high agreement),\\
1: High disagreement (low agreement).\\
Format your response as follows:\\
``answer: \{score\}; explanation: \{reasoning\}''}
\paragraph{Pairwise} 
\textit{Here are two images of advertisement. Which one is more abnormal and unusual? 1 for the left image and 2 for the right image. Give your answer in the following format: ``answer: \{number\}; explanation: \{reasoning\}''}

\section{Output Analysis}
\label{sec:appendix_output_analysis}
\subsection{Error Analysis on Pairwise Outputs}
\label{sec:appendix_error_analysis}
\begin{figure*}
    \centering
    \begin{minipage}[b]{0.45\textwidth}
        \includegraphics[width=\textwidth]{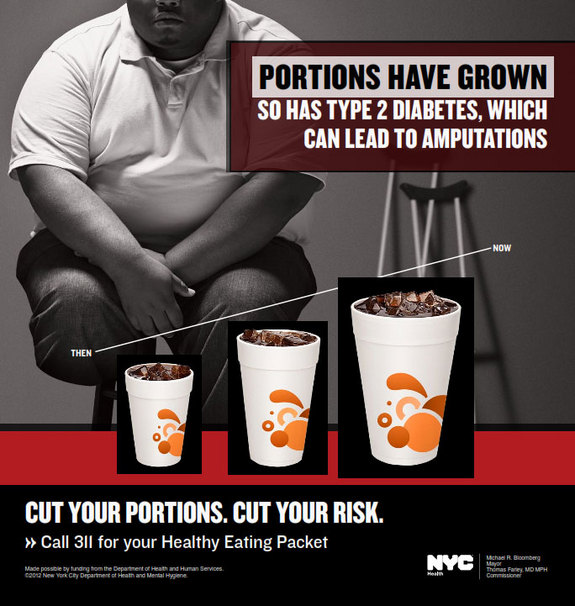}
        % \caption{Flower one.}
    \end{minipage}
    \hfill
    \begin{minipage}[b]{0.45\textwidth}
        \includegraphics[width=\textwidth]{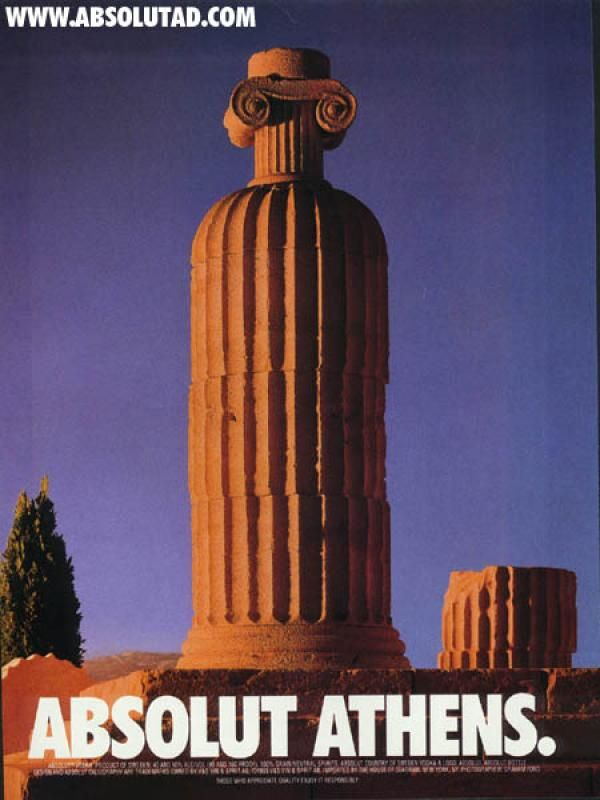}
    \end{minipage}
    \caption{Task: \textbf{atypicality}; model outputs are in Table \ref{table:error_analysis_1}}
    \label{fig:error_analysis_1_img}
\end{figure*}

\begin{figure*}
    \centering
    \begin{minipage}[b]{0.45\textwidth}
        \includegraphics[width=\textwidth]{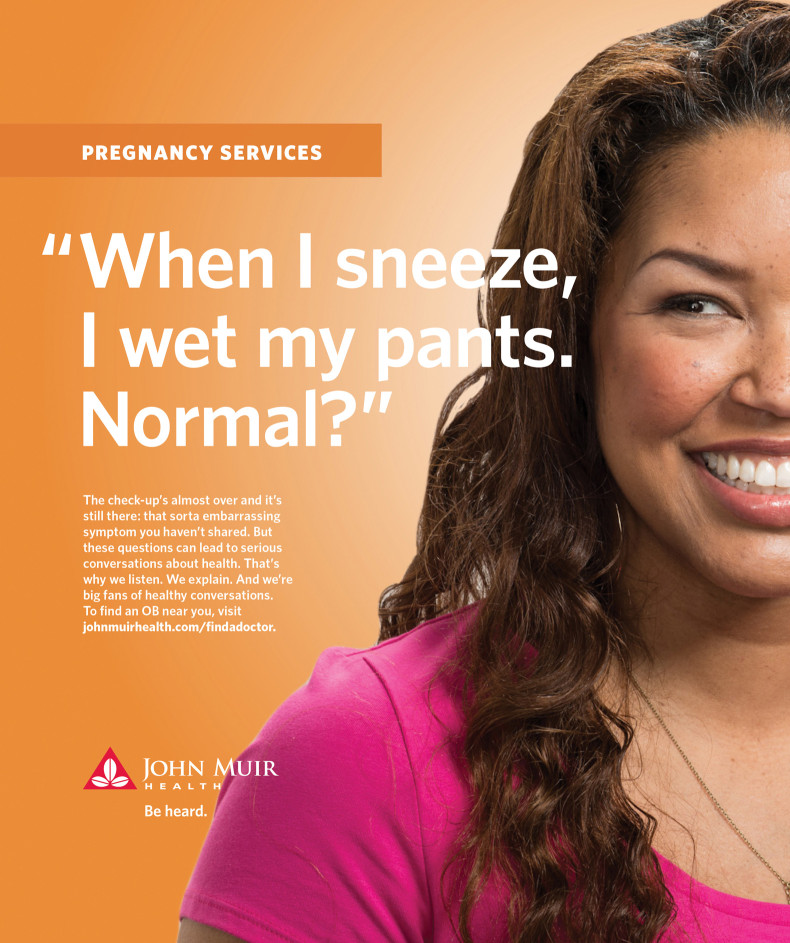}
        % \caption{Flower one.}
    \end{minipage}
    \hfill
    \begin{minipage}[b]{0.45\textwidth}
        \includegraphics[width=\textwidth]{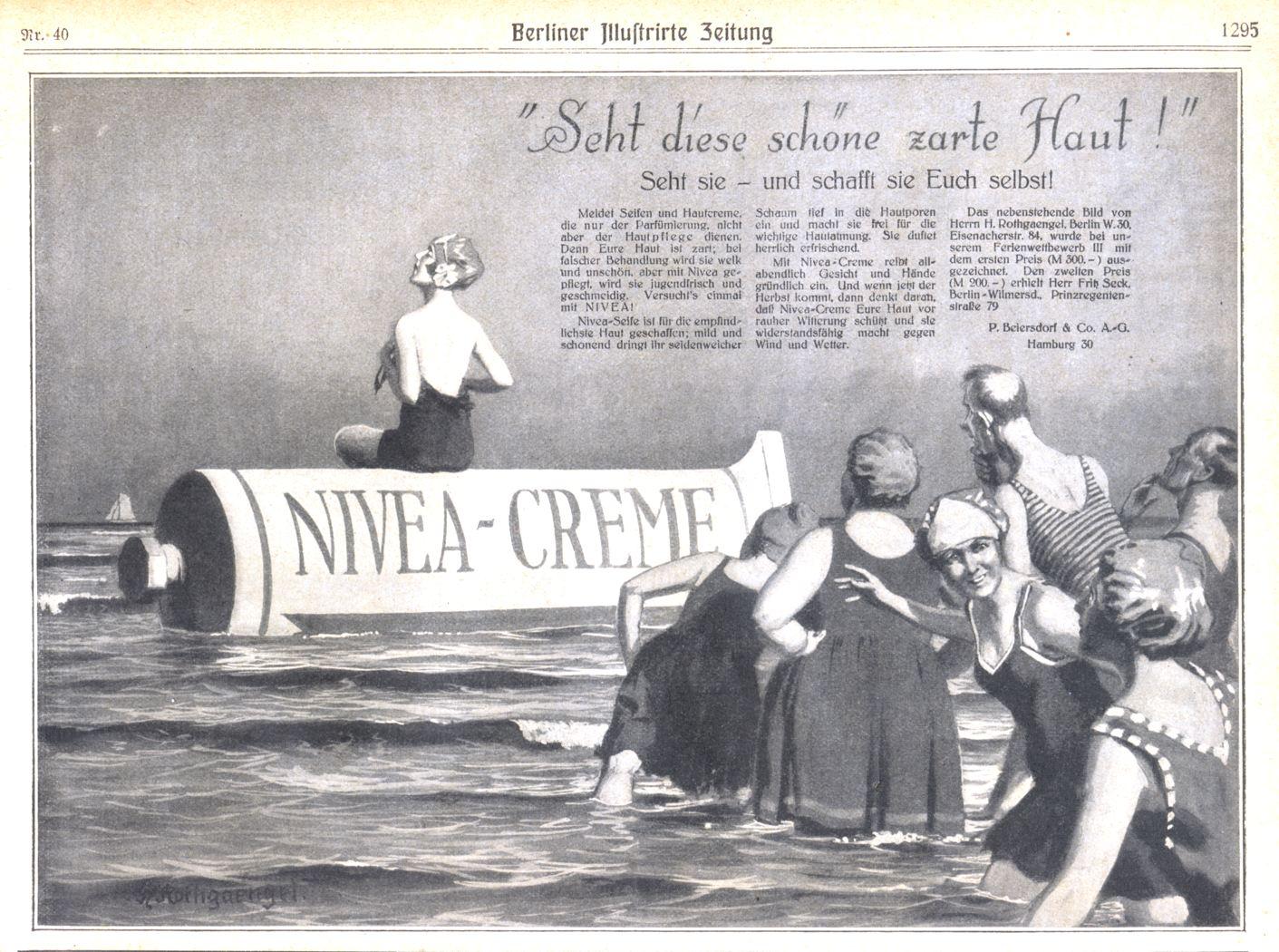}
    \end{minipage}
    \caption{Task: \textbf{originality}; model outputs are in Table \ref{table:error_analysis_2}}
    \label{fig:error_analysis_2_img}
\end{figure*}

Here, we present two error analysis examples on the pairwise task. Detailed analysis are in the caption of each image pair (Figure \ref{fig:error_analysis_1_img}, \ref{fig:error_analysis_2_img}).

%%%%%%%%%%%%%%%%%%%%%% Error Analysis 1 %%%%%%%%%%%%%%%%%%%%%%%%%%%%%%%%%%%%%%%%%%%%%%%

\begin{table*}
\small
\centering
    \begin{tabular}{p{0.12\textwidth} p{0.71\textwidth} p{0.09\textwidth}}
        \toprule
        \textbf{Model} & \textbf{Model Output}  & \textbf{Prediction} \\
        \midrule
        LLaVA-7B & explanation: The image on the left features a man sitting on a chair with a cup of soda in front of him. The text on the ad is about the risks of consuming sugary drinks and the importance of cutting one's portion sizes. This is a typical advertisement that uses a real-life scenario to convey a message. On the other hand, the image on the right is more unusual and abnormal. It features a large, brown, cylindrical structure that resembles a bottle or a vase. The text on this ad is about the risks of consuming sugary drinks and the importance of cutting one's portion sizes. The presence of the large structure in the ad is unexpected and out of place, making it more unusual and abnormal compared to the first image. answer: 2 & right (2) \\
        \midrule
        LLaVA-13B & The image on the left is more abnormal and unusual because it features a person with a significantly larger waistline than the rest of their body, which is not a typical human proportion. \textbf{The image on the right shows a large, ancient-looking structure, which is not an unusual sight in itself, but it is presented in a way that suggests it is a part of the advertisement, which is not a common practice for advertising.} & left (1)\\
        \midrule
        InternVL2-8B & explanation: The left advertisement features a person with a large belly, which is a common depiction in media to represent obesity. The use of a person in the image is typical in health campaigns to visually convey the message about portion sizes and health risks. The right advertisement, on the other hand, uses the Parthenon in Athens, an ancient monument, to convey the message about portion sizes. Using a historical or cultural landmark in an advertisement is less common and more unusual compared to using a person. Therefore, the right advertisement is more abnormal and unusual. answer: 2 \textbf{\textit{\{repeating answer 4 times\}}} & right (2) \\
        \midrule
        GPT-4v & explanation: The advertisement on the right is more abnormal and unusual because it portrays a well-known architectural column, which is typically straight, as a bottle of Absolut Vodka. This is a surreal and artistic reinterpretation of a cultural landmark, intended to resemble the shape of the iconic Absolut bottle, which is not a normal sight in real-world settings. On the other hand, the advertisement on the left, while addressing a serious health issue regarding portion sizes and diabetes, doesn't alter common objects in an unexpected way but uses real-life elements (a person and soft drink cups) to convey its message. Therefore, the right image is more surreal and unconventional in its approach. answer: 2 & right (2)\\
        \bottomrule
    \end{tabular}
    \caption{Model output for ads pair in Figure \ref{fig:error_analysis_1_img}. The correct answer is right (2). We can see LLaVA-13B is the only model that gets the answer incorrectly and its explanation is also self-contradicting (the bold part); for InternVL2, the problem is that it always repeat the answer for multiple times, even when it answers correctly.}
    \label{table:error_analysis_1}
\end{table*}

%%%%%%%%%%%%%%%%%%%%%% Error Analysis 2 %%%%%%%%%%%%%%%%%%%%%%%%%%%%%%%%%%%%%%%%%%%%%%%

\begin{table*}
\small
\centering
    \begin{tabular}{p{0.12\textwidth} p{0.71\textwidth} p{0.09\textwidth}}
        \toprule
        \textbf{Model} & \textbf{Model Output}  & \textbf{Prediction} \\
        \midrule
        LLaVA-7B & explanation: The left image features a woman with a caption that reads ``When I sneeze, I wet my pants. Normal?'' This ad is unique because it addresses a common but often embarrassing issue, \textbf{using humor to connect with the audience}. The right image, on the other hand, shows a group of women in a bathing suit, with a caption that reads: \textbf{Seh die Schönheit! Seh die Schönheit! Seh die Schönheit! Seh die Schönheit! Seh die Schönheit! Seh die Schönheit! Seh die Schönheit!} \textbf{\textit{\{repeat lot of times\}}} & left (1) \\
        \midrule
        LLaVA-13B & The left image is more unique compared to the right one. The reason is that the left image features a modern woman with a contemporary hairstyle and clothing, while the right image is an old-fashioned advertisement with a vintage design and illustration style. \textbf{The juxtaposition of the modern woman with the vintage advertisement creates a contrast that makes the left image more unique.} & left (1) \\
        \midrule
        InternVL2-8B & explanation: The left advertisement is more unique compared to the right one because it uses a modern approach with a relatable and humorous question about a common pregnancy symptom, which can engage and resonate with potential customers. The use of a smiling woman and a direct question makes it more approachable and relevant to current audiences. In contrast, the right advertisement uses a vintage style and focuses on a specific product, which might not immediately connect with modern viewers as well. \textbf{The humor and relatability in the left ad make it stand out in the same product category}. Answer: 1 \textbf{\textit{\{repeating answer 4 times\}}} & left (1) \\
        \midrule
        GPT-4v & explanation: The advertisement on the left provides a candid, contemporary approach to discussing health issues, specifically addressing pregnancy-related concerns in a way that openly discusses a commonly experienced, yet often unspoken issue. This upfront and direct style can be considered unique in comparison to typical pregnancy services ads which may traditionally use softer language and imagery. On the right, the \textbf{Nivea advertisement} takes a more classic approach, using vintage-style artwork to represent the product in a historical context, \textbf{which is creative but not necessarily unique as it aligns with a familiar tradition of period-specific advertising.} This style, though distinctive, may not stand out as much in the skincare product category where vintage and nostalgic themes are frequently employed. answer: 1 &  left (1) \\
        \bottomrule
    \end{tabular}
    \caption{Model output for ads pair in Figure \ref{fig:error_analysis_2_img}. The correct answer is right (2). ALLaVA-13B is the only one that incorrectly combined the visual element in the left image (``modern woman'') and the right image (``vintage advertisement'') together as if they are in the same ad (``the juxtaposition of...''); LLaVA-7B and InternVL2-8B are able to capture the humor in the left ad but still make the mistake. In particular, LLaVA-7B is stuck in consistently generating ``Seh die Schönheit!'', which is a description of the left image, without even get a chance to get to the second image. This might be the reason for its failure. Note that GPT-4v is the only model that recognizes the brand name (``Nivea''), highlighting its strong capability in recognizing details.}
    \label{table:error_analysis_2}
\end{table*}

\subsection{Distribution Modeling Examples}
\label{sec:appendix_distribution_output}
\begin{figure*}
    \centering
    \begin{minipage}[b]{0.45\textwidth}
        \includegraphics[width=\textwidth]{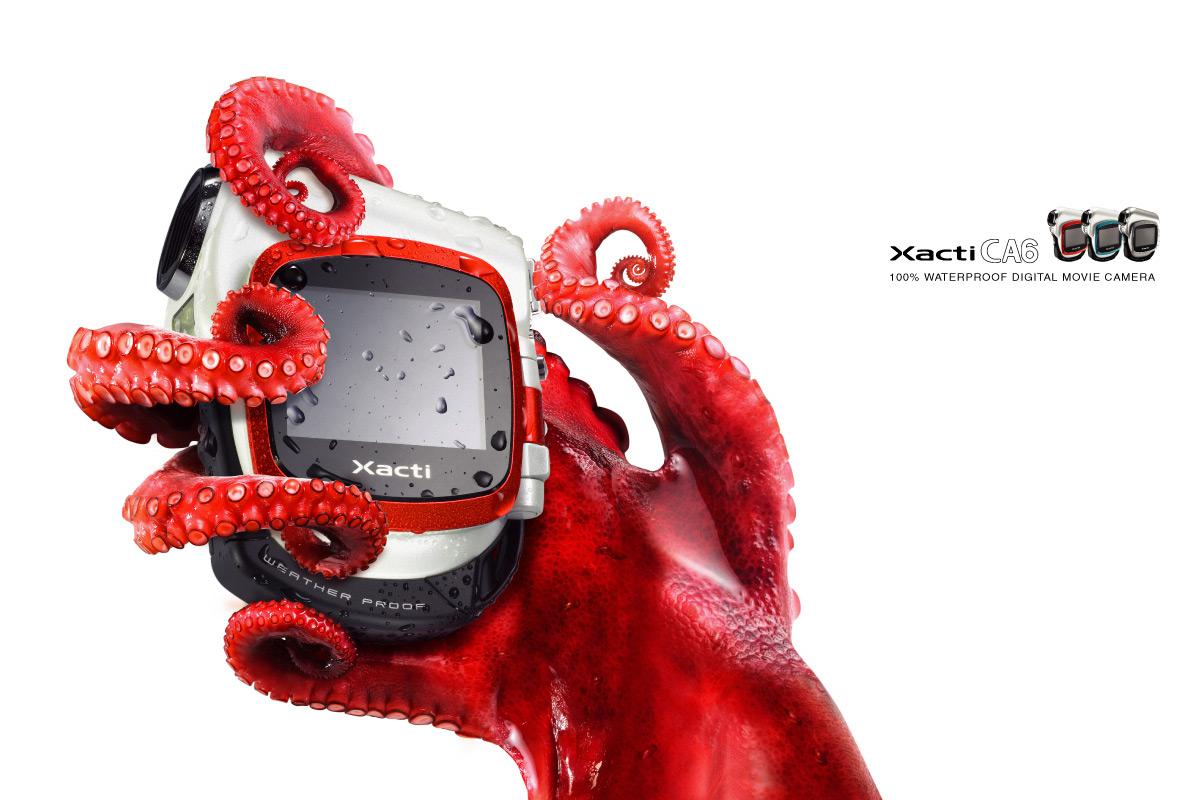}
        % \caption{Flower one.}
    \end{minipage}
    \hfill
    \begin{minipage}[b]{0.45\textwidth}
        \includegraphics[width=\textwidth]{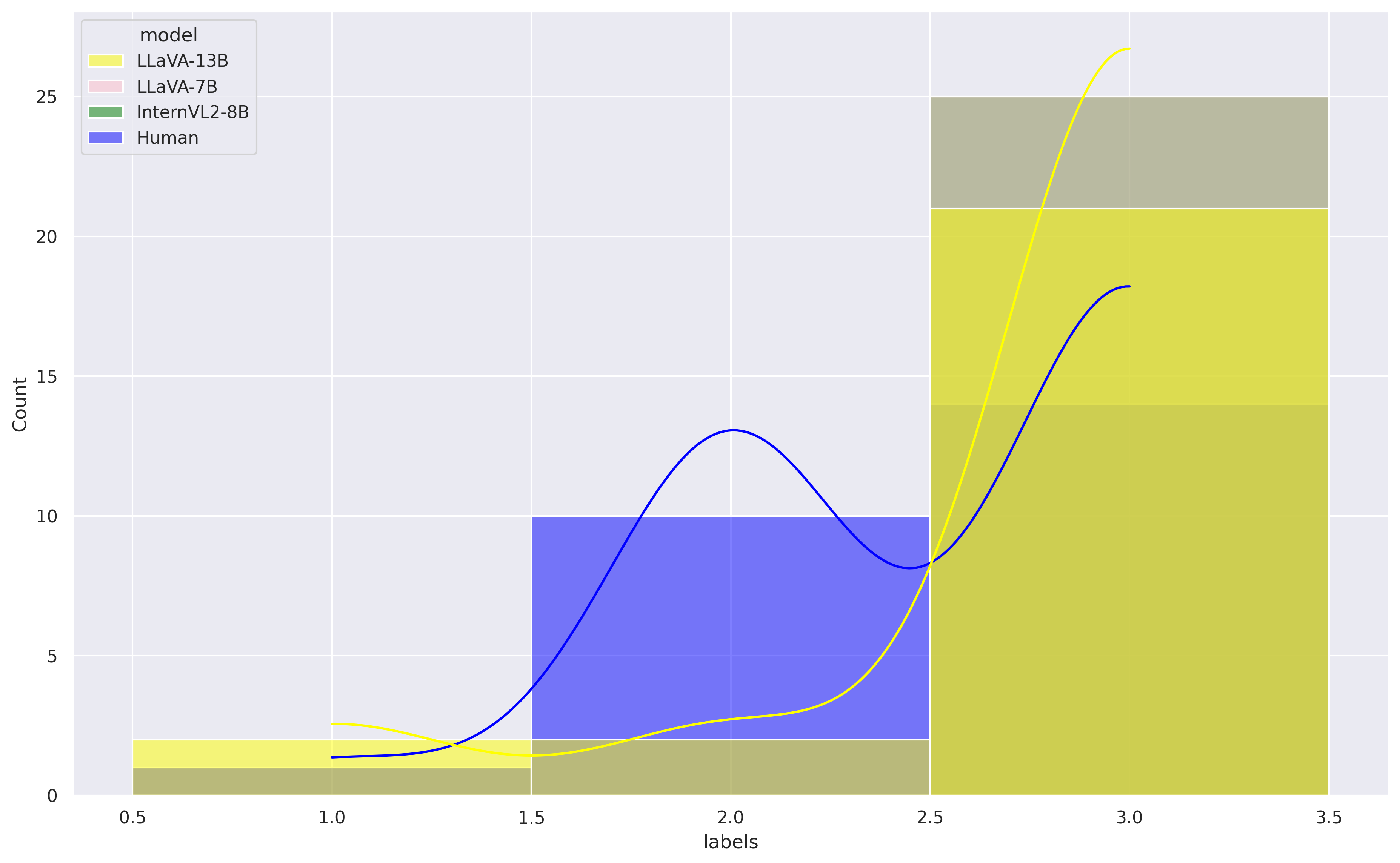}
    \end{minipage}
    \caption{Example (A) and \texttt{creativity} predictions by models; complete output in Table \ref{table:output_example_1}}
    \label{fig:output_example_1}
\end{figure*}

\begin{figure*}
    \centering
    \begin{minipage}[b]{0.45\textwidth}
        \includegraphics[width=\textwidth]{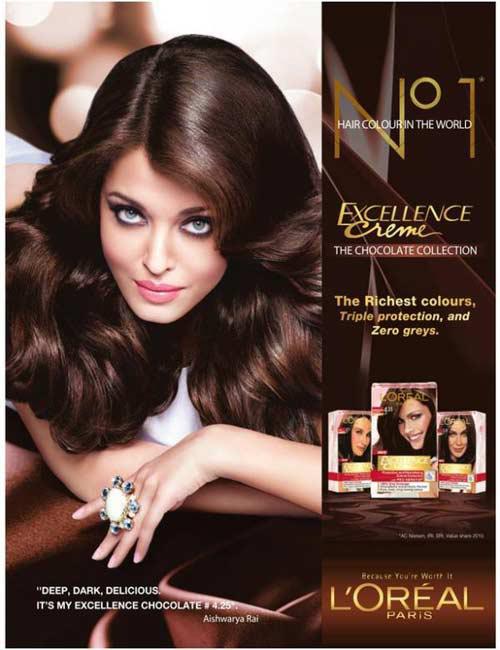}
        % \caption{Flower one.}
    \end{minipage}
    \hfill
    \begin{minipage}[b]{0.45\textwidth}
        \includegraphics[width=\textwidth]{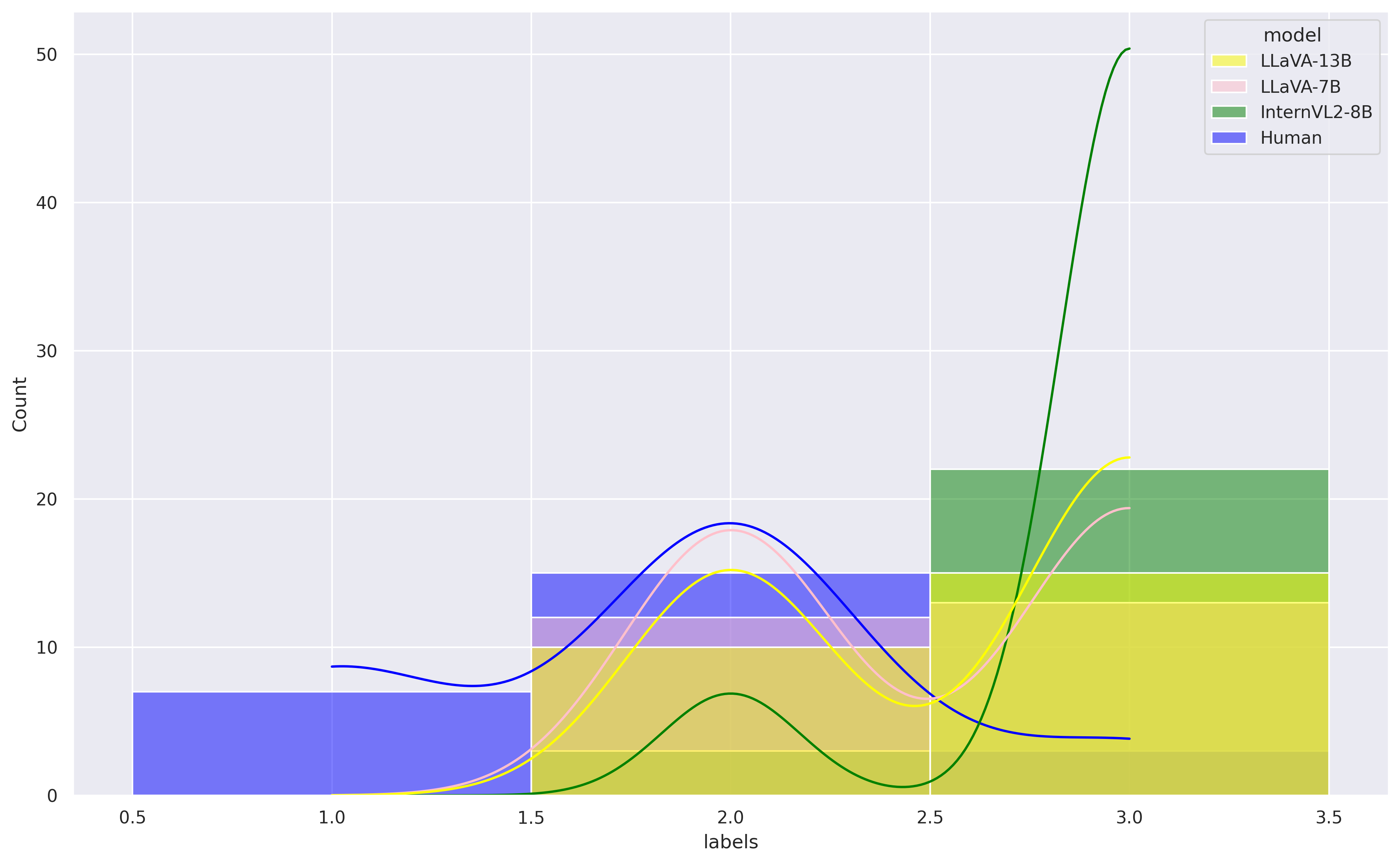}
    \end{minipage}
    \caption{Example (B) and \texttt{creativity} predictions by models; complete output in Table \ref{table:output_example_2}}
    \label{fig:output_example_2}
\end{figure*}

\begin{figure*}
    \centering
    \begin{minipage}[b]{0.45\textwidth}
        \includegraphics[width=\textwidth]{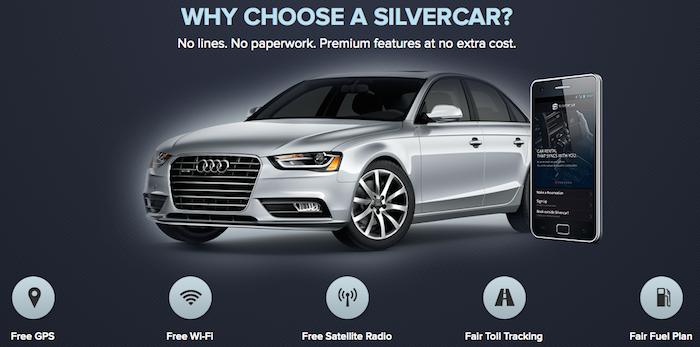}
        % \caption{Flower one.}
    \end{minipage}
    \hfill
    \begin{minipage}[b]{0.45\textwidth}
        \includegraphics[width=\textwidth]{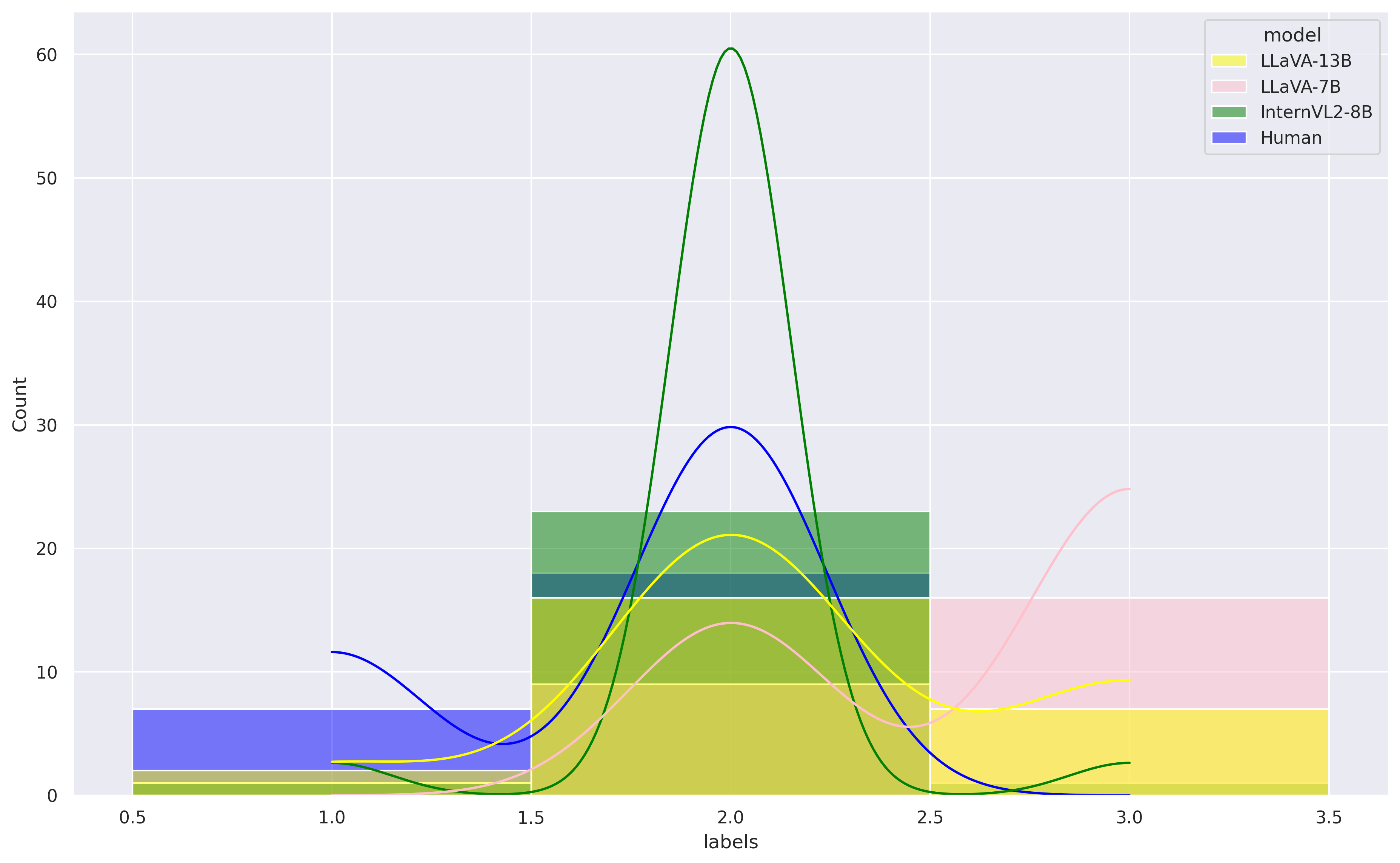}
    \end{minipage}
    \caption{Example (C) and \texttt{creativity} predictions by models; complete output in Table \ref{table:output_example_3}}
    \label{fig:output_example_3}
\end{figure*}

\begin{figure*}
    \centering
    
    \begin{minipage}[b]{0.25\textwidth}
        \includegraphics[width=\textwidth]{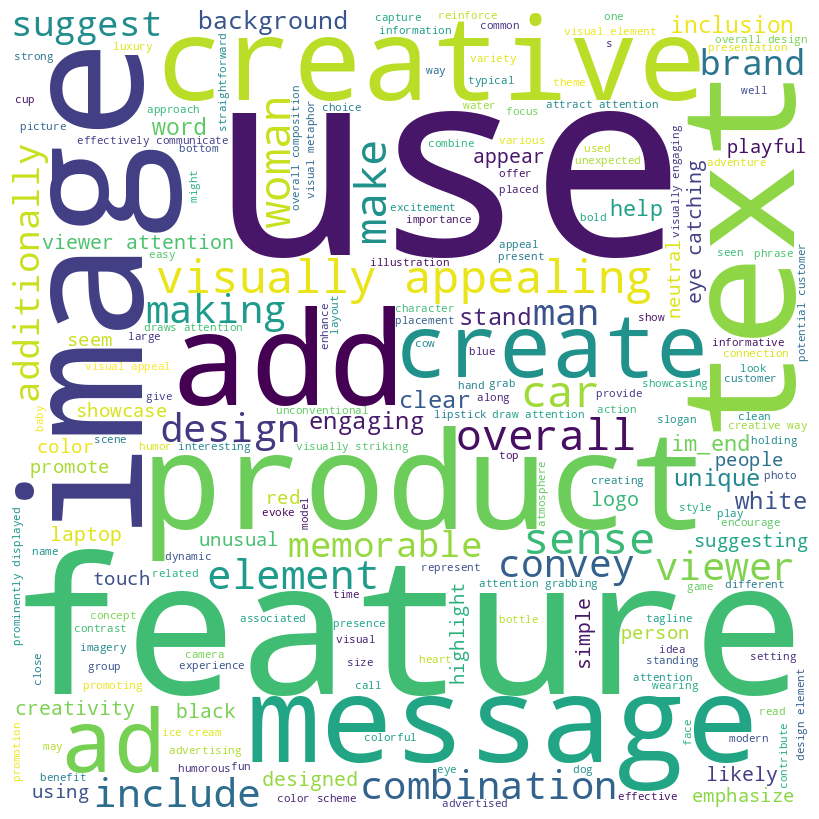}
        \caption{Task: Creativity, Model: LLaVA-7B}
    \end{minipage}
    \hfill
    \begin{minipage}[b]{0.25\textwidth}
        \includegraphics[width=\textwidth]{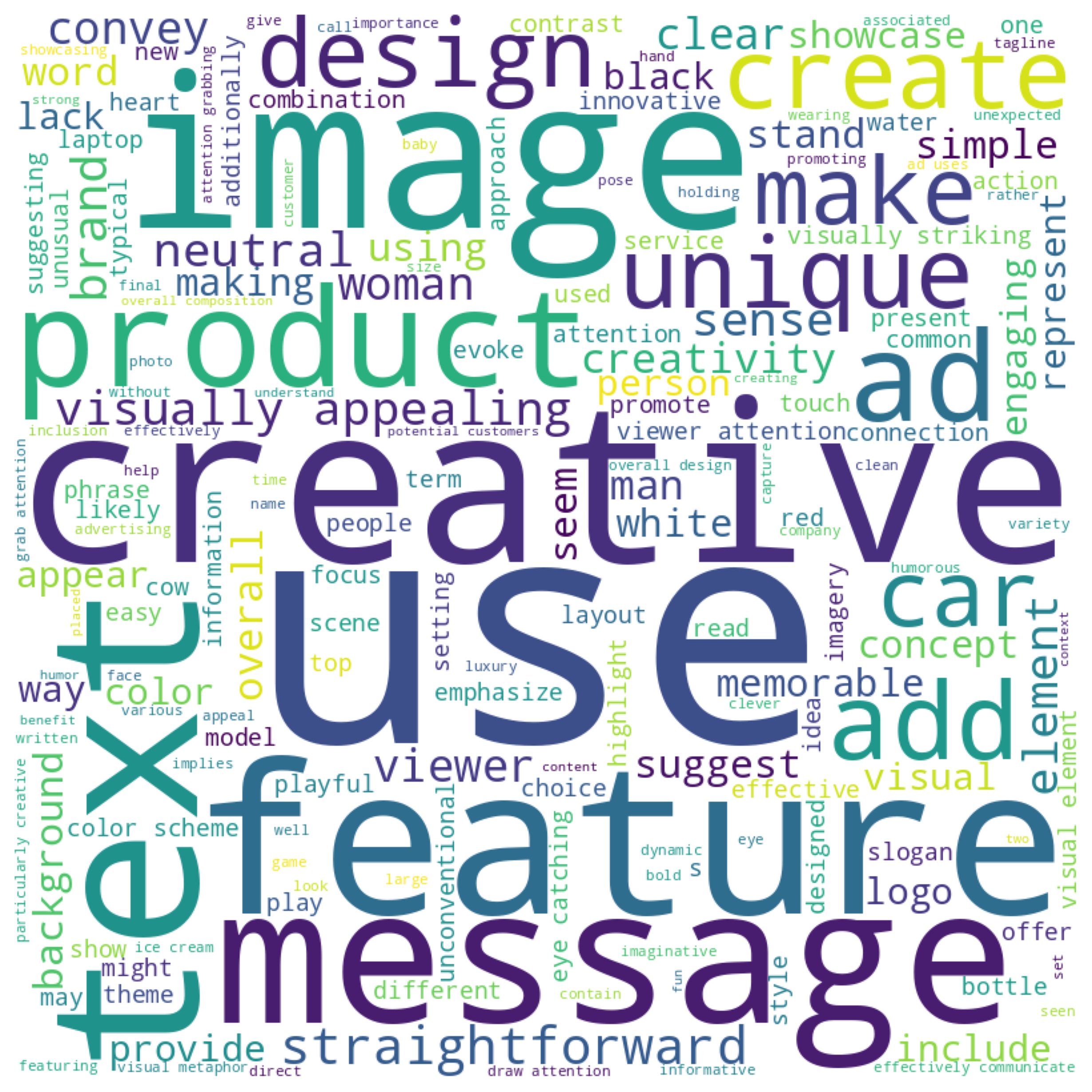}
        \caption{Task: Creativity, Model: LLaVA-13B}
    \end{minipage}
    \hfill
    \begin{minipage}[b]{0.25\textwidth}
        \includegraphics[width=\textwidth]{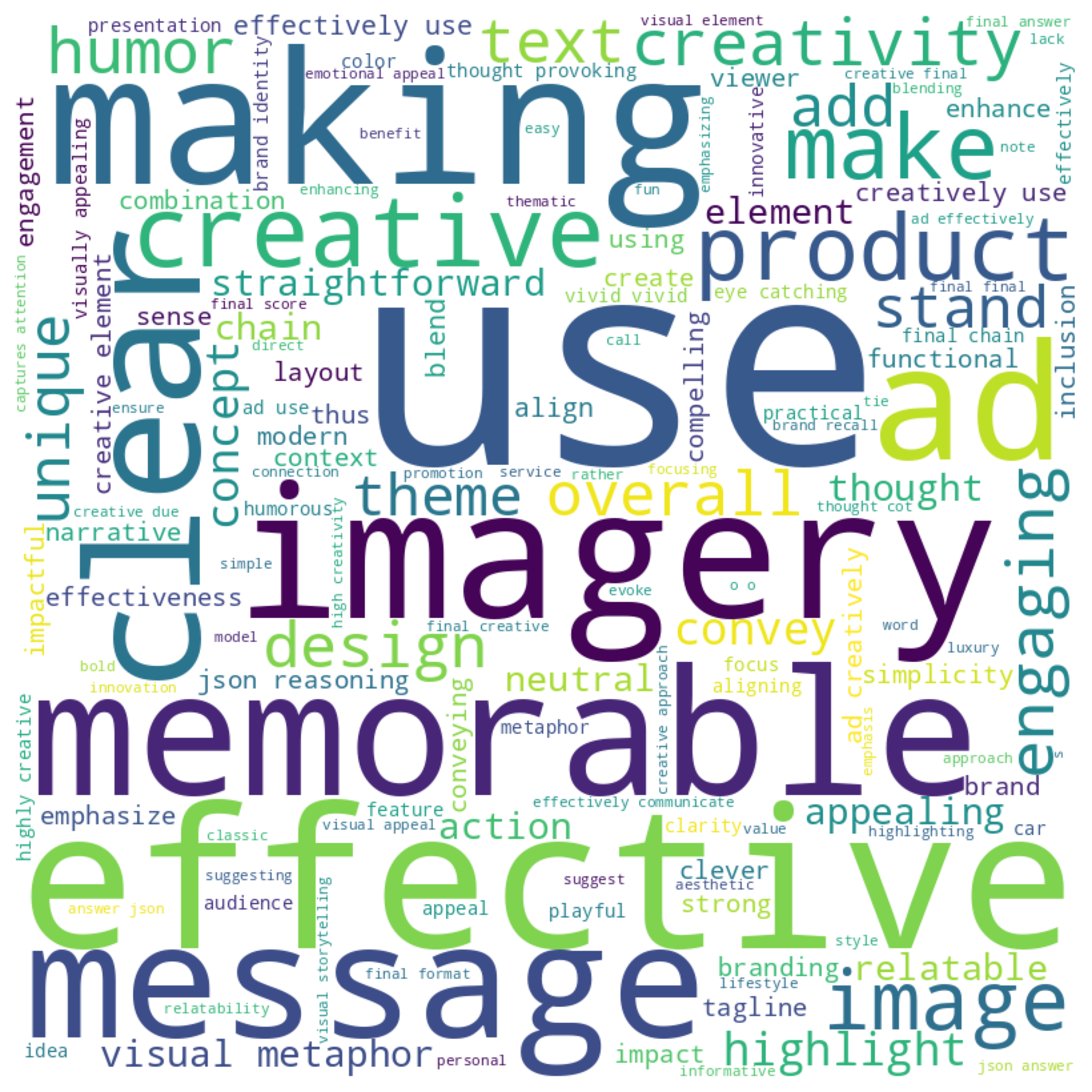}
        \caption{Task: Creativity, Model: InternVL2-8B}
        \label{fig:wordcloud_1}
    \end{minipage}
    
\end{figure*}

\begin{figure*}
    \centering
    \begin{minipage}[b]{0.25\textwidth}
        \includegraphics[width=\textwidth]{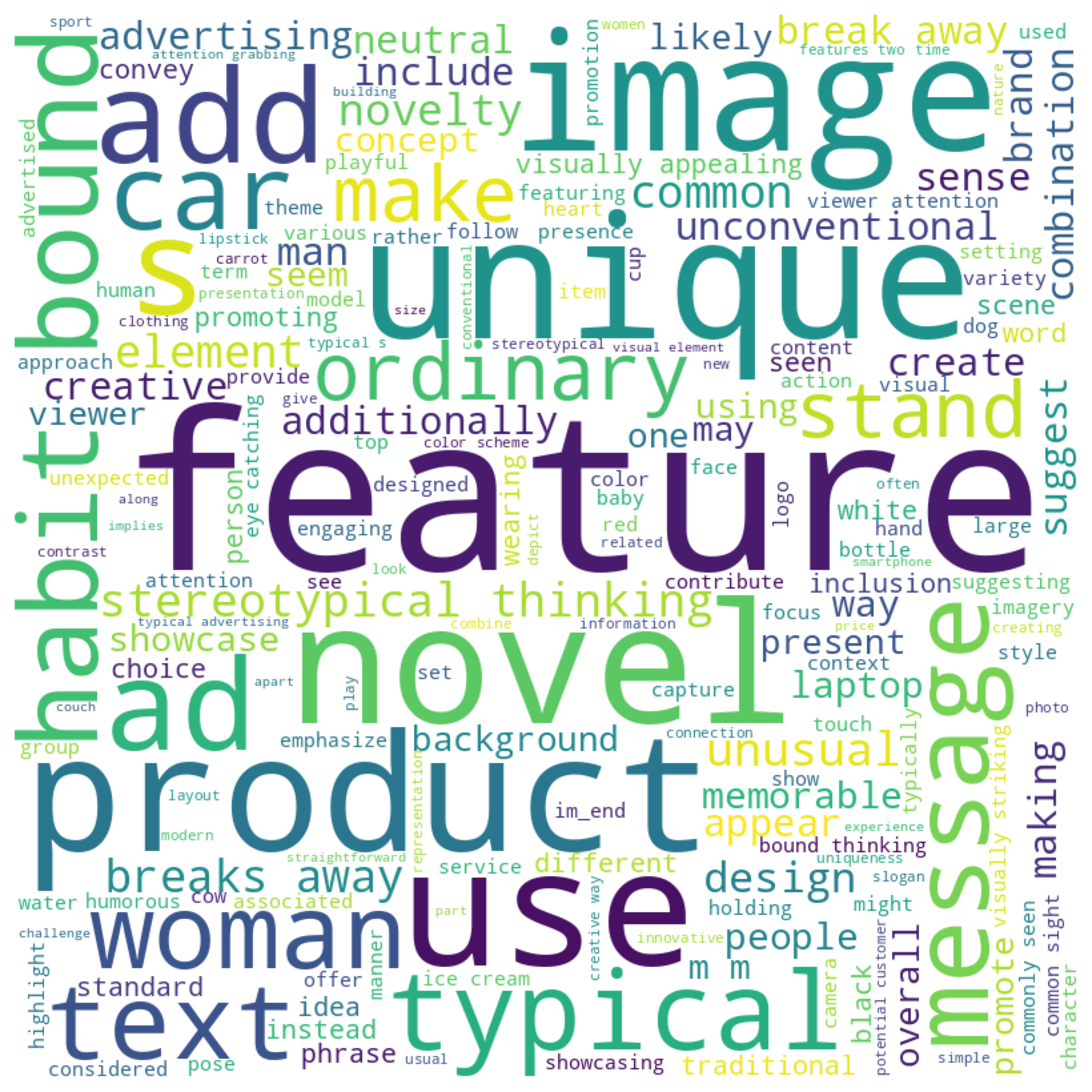}
        \caption{Task: Originality, Model: LLaVA-7B}
    \end{minipage}
    \hfill
    \begin{minipage}[b]{0.25\textwidth}
        \includegraphics[width=\textwidth]{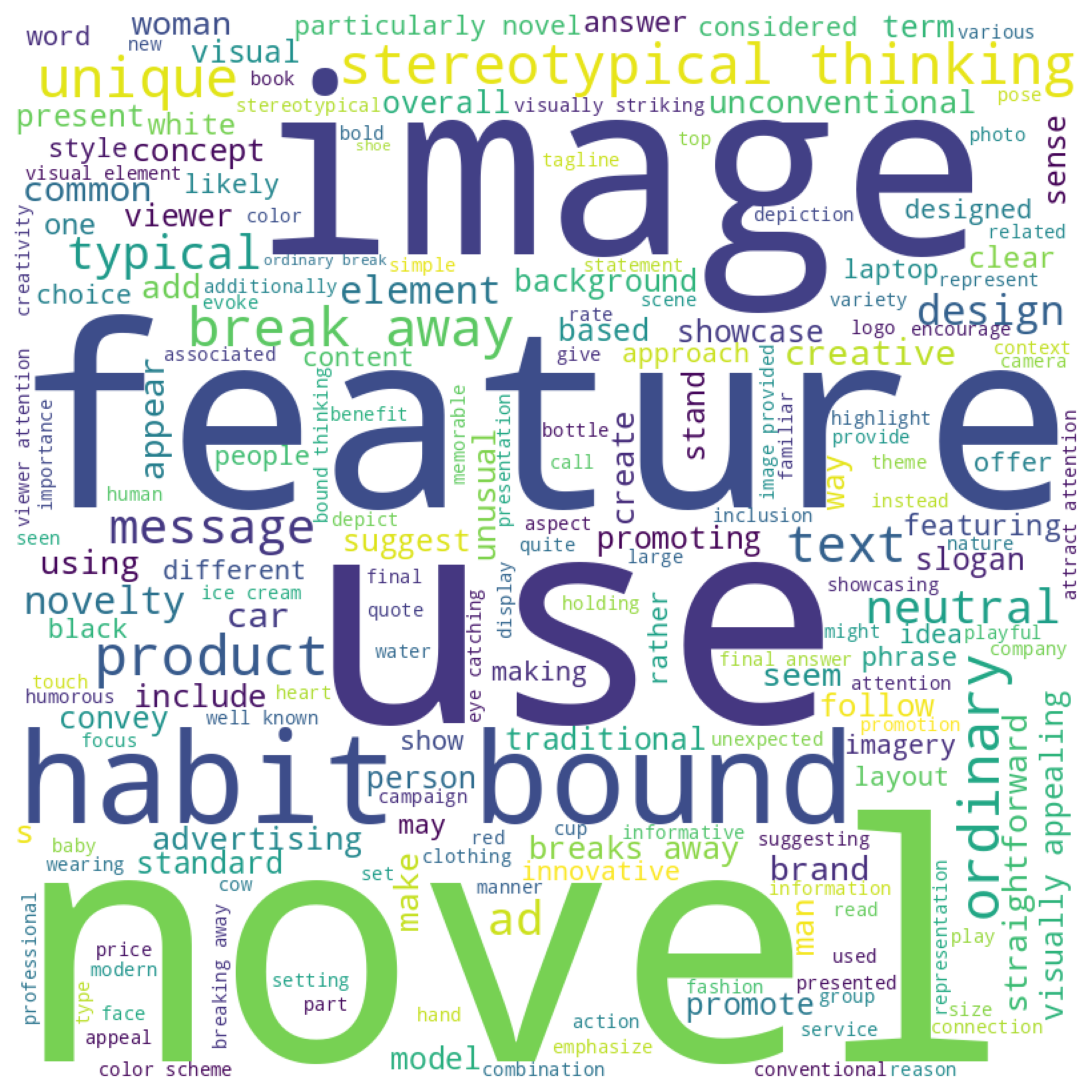}
        \caption{Task: Originality, Model: LLaVA-13B}
    \end{minipage}
    \hfill
    \begin{minipage}[b]{0.25\textwidth}
        \includegraphics[width=\textwidth]{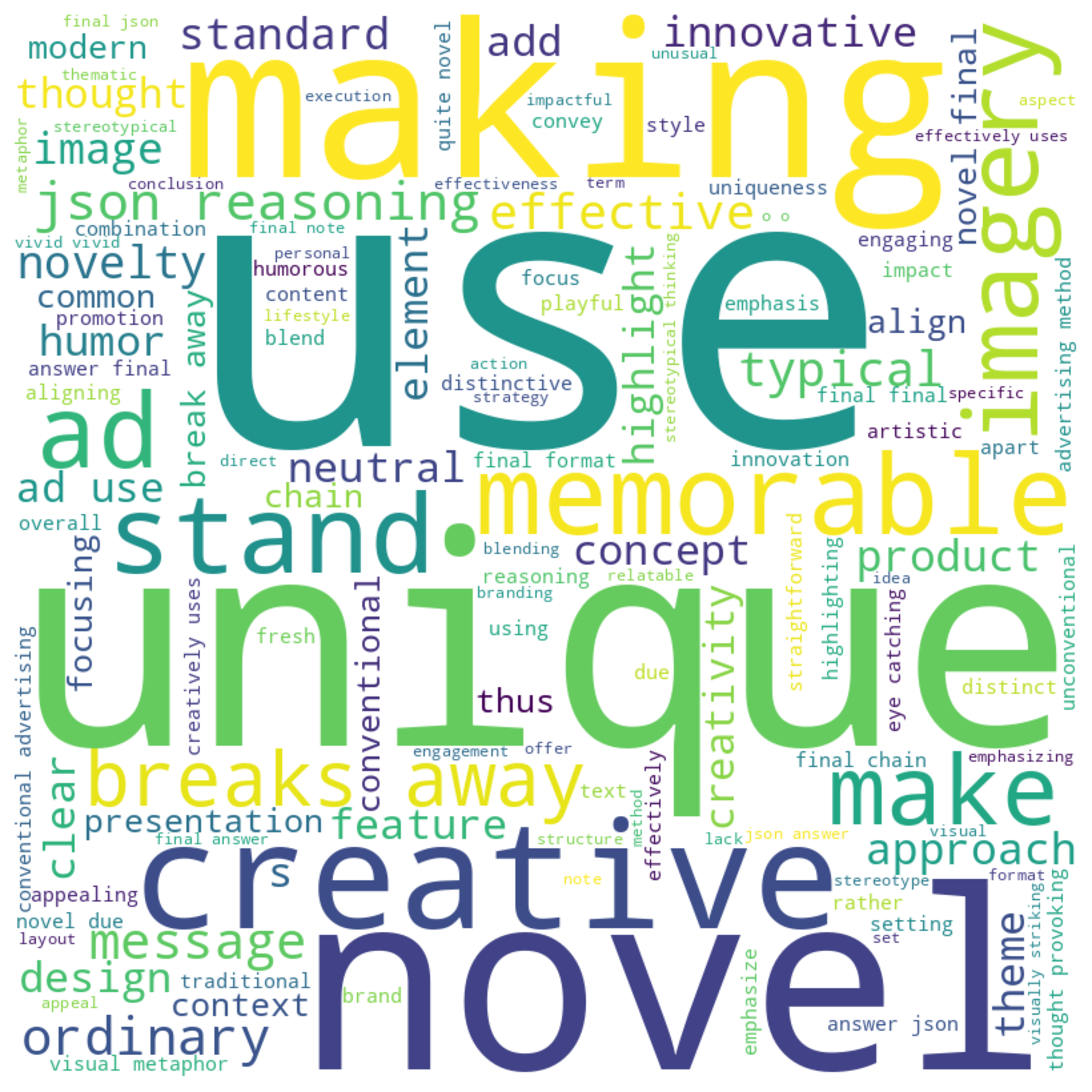}
        \caption{Task: Originality, Model: InternVL2-8B}
        \label{fig:wordcloud_2}
    \end{minipage}
    
\end{figure*}

\begin{figure*}
    \centering
    \begin{minipage}[b]{0.25\textwidth}
        \includegraphics[width=\textwidth]{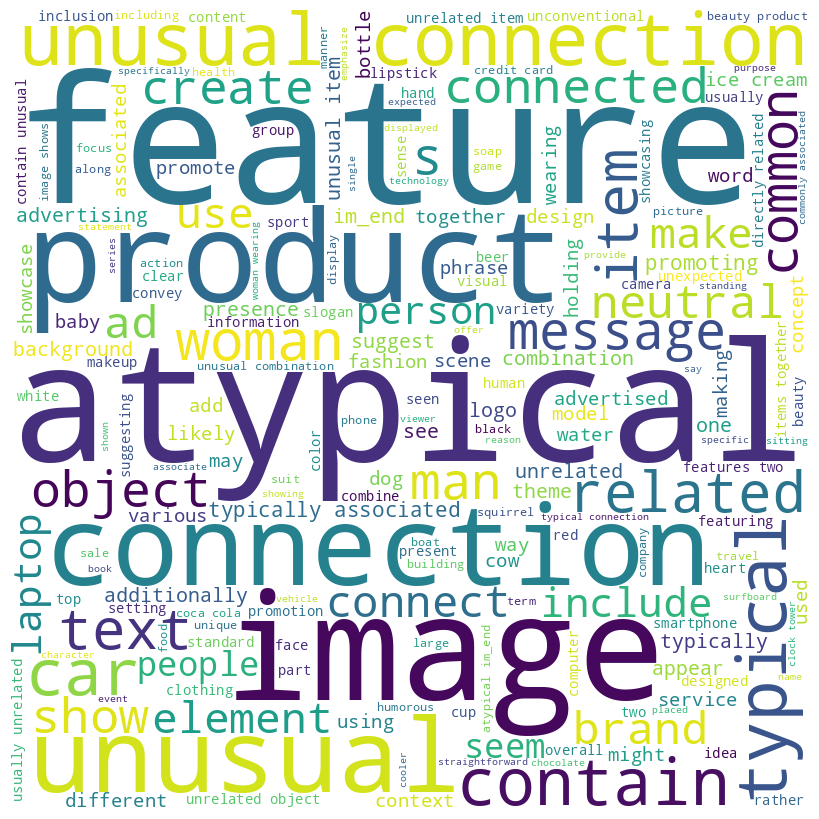}
        \caption{Task: Atypicality, Model: LLaVA-7B}
    \end{minipage}
    \hfill
    \begin{minipage}[b]{0.25\textwidth}
        \includegraphics[width=\textwidth]{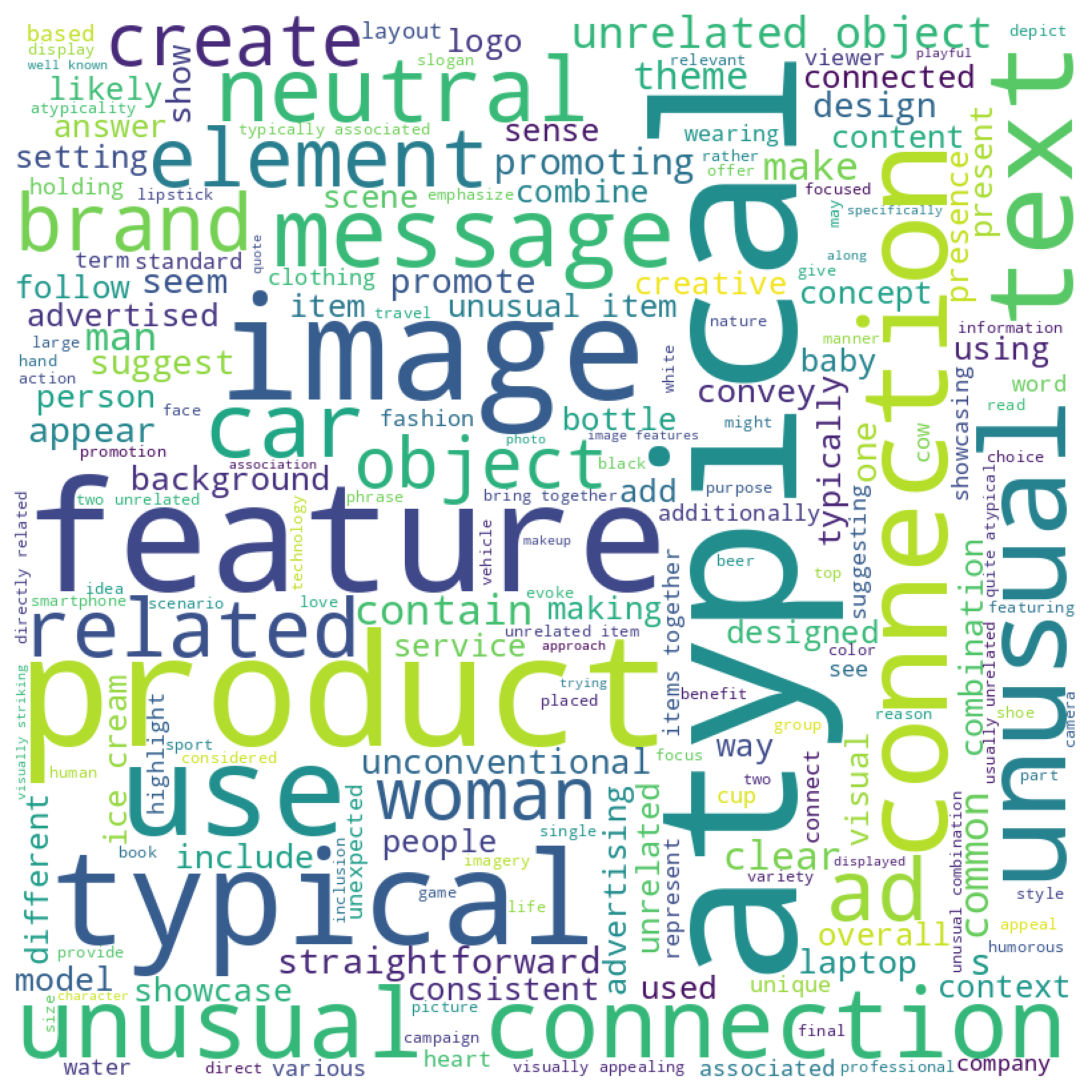}
        \caption{Task: Atypicality, Model: LLaVA-13B}
    \end{minipage}
    \hfill
    \begin{minipage}[b]{0.25\textwidth}
        \includegraphics[width=\textwidth]{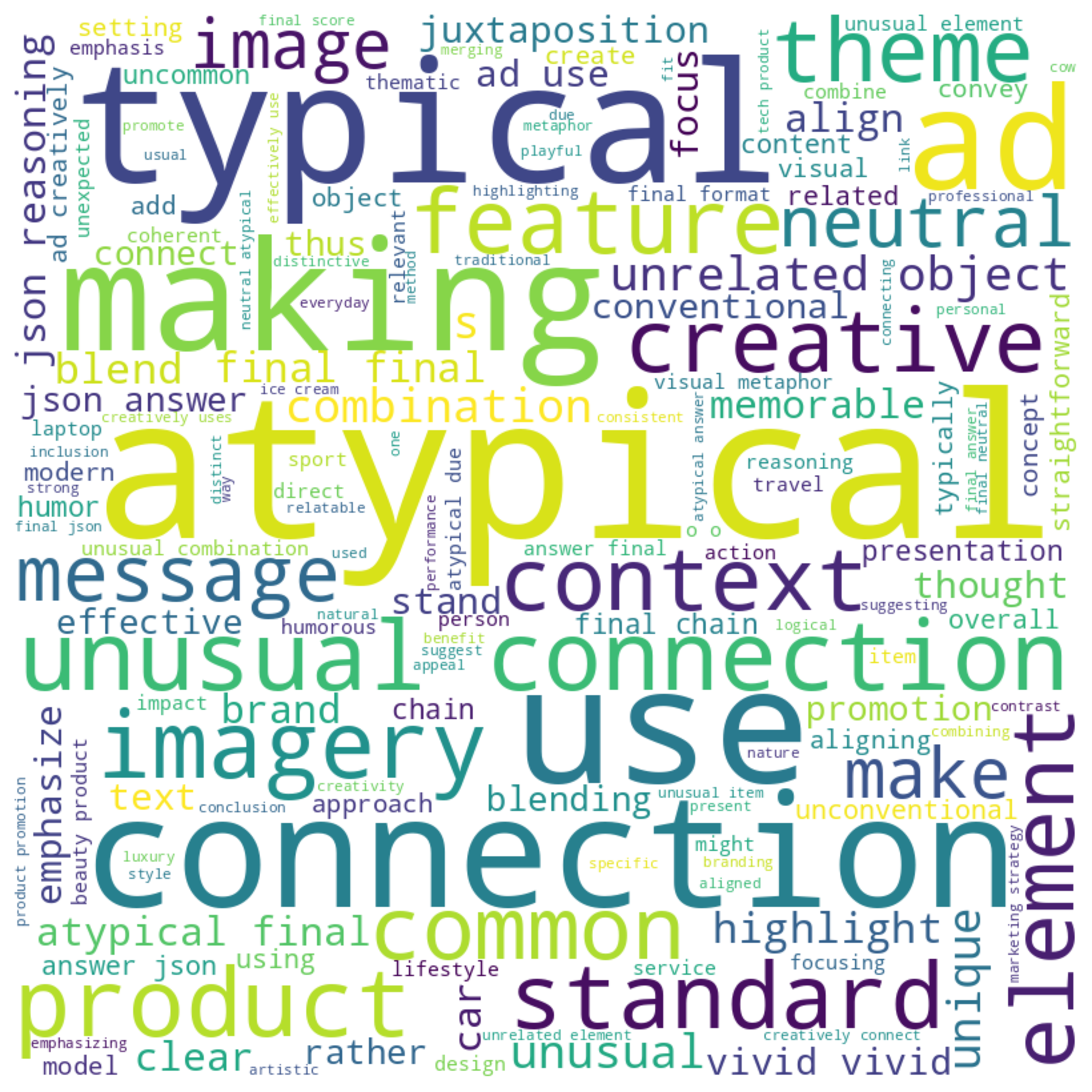}
        \caption{Task: Atypicality, Model: InternVL2-8B}
        \label{fig:wordcloud_3}
    \end{minipage}
    
\end{figure*}

We have three examples with all the scoring metrics; see Figure \ref{fig:output_example_1}, \ref{fig:output_example_2}, \ref{fig:output_example_3}. We have also plotted WordClouds for the \textit{reasoning} part of the output (Figure \ref{fig:wordcloud_1}, \ref{fig:wordcloud_2}, \ref{fig:wordcloud_3}). As shown in those word clouds, the commonly used phrases generally closely correspond to the task definition:  `creative', `message', `imagery', `effective' for creativity task, `unique', `novel', `habit' for originality task, and `atypical', `connection', `
unusual' for atypicality task).

%%%%%%%%%%%%%%%%%%%%%%%%%%%%%%%%%%%%%%%%%%%%%%%%%%%%%%%%%%%%%%%%%%%%%%%%%%%%%%%%%%%%%%%%%%%%%%%

\begin{table*}
\small
\centering
    \begin{tabular}{ccccc}
        \toprule
        \textbf{Aspect} & \textbf{Human} & \textbf{LLaVA-7B \textit{(KL)}} & \textbf{LLaVA-13B \textit{(KL)}}& \textbf{InternVL \textit{(KL)}}  \\
        \midrule
        Creativity & 2.60 & 3.00 (1.0054) & 2.76 (0.3986) & 3.00 (1.0054) \\
        \midrule
        Originality & 2.92 & 2.92 (0.9643) & 2.88 (0.3144) & 3.00 (1.0054)  \\
         \midrule
        Atypicality & 2.92 & 2.88 (0.6030) & 2.64 (0.1191) & 2.84 (0.2223) \\
        \bottomrule
    \end{tabular}
    \caption{Model output (average across 25 runs) and human ratings for Example (A), see ad image and distribution modeling result in Figure \ref{fig:output_example_1}; \textit{KL} refers to $KL(Human||Model)$}
    \label{table:output_example_1}
\end{table*}

%%%%%%%%%%%%%%%%%%%%%%%%%%%%%%%%%%%%%%%%%%%%%%%%%%%%%%%%%%%%%%%%%%%%%%%%%%%%%%%%%%%%%%%%%%%%%%%

\begin{table*}
\small
\centering
    \begin{tabular}{ccccc}
        \toprule
        \textbf{Aspect} & \textbf{Human} & \textbf{LLaVA-7B \textit{(KL)}} & \textbf{LLaVA-13B \textit{(KL)}}& \textbf{InternVL \textit{(KL)}}  \\
        \midrule
        Creativity & 2.60 & 2.52 (0.7701) & 2.60 (0.8803) & 2.88 (1.6395) \\
        \midrule
        Originality & 2.92 & 2.28 (0.2762) & 1.60 (0.0963) & 2.12 (0.1791) \\
         \midrule
        Atypicality & 2.92 & 1.76 (0.2560) & 1.68 (0.070) & 1.36 (0.5113) \\
        \bottomrule
    \end{tabular}
    \caption{Model output and human ratings for Example (B), see ad image and distribution modeling result in Figure \ref{fig:output_example_2}; \textit{KL} refers to $KL(Human||Model)$}
    % \vspace{-6mm}
    \label{table:output_example_2}
\end{table*}

%%%%%%%%%%%%%%%%%%%%%%%%%%%%%%%%%%%%%%%%%%%%%%%%%%%%%%%%%%%%%%%%%%%%%%%%%%%%%%%%%%%%%%%%%%%%%%%
% \newpage

\begin{table*}
\small
\centering
    \begin{tabular}{ccccc}
        \toprule
        \textbf{Aspect} & \textbf{Human} & \textbf{LLaVA-7B \textit{(KL)}} & \textbf{LLaVA-13B \textit{(KL)}}& \textbf{InternVL \textit{(KL)}}  \\
        \midrule
        Creativity & 2.60 & 2.64 (1.3395) & 2.20 (0.4060) & 2.00 (0.3070) \\
        \midrule
        Originality & 2.92 & 2.36 (0.7893) & 1.56 (0.0643) & 1.60 (0.3263) \\
         \midrule
        Atypicality & 2.92 & 1.80 (0.6703) & 1.60 (0.5048) & 1.16 (0.8899) \\
        \bottomrule
    \end{tabular}
    \caption{Model output and human ratings for Example (C), see ad image and distribution modeling result in Figure \ref{fig:output_example_3}; \textit{KL} refers to $KL(Human||Model)$}
    % \vspace{-6mm}
    \label{table:output_example_3}
\end{table*}

%%%%%%%%%%%%%%%%%%%%%%%%%%%%%%%%%%%%%%%%%%%%%%%%%%%%%%%%%%%%%%%%%%%%%%%%%%%%%%%%%%%%%%%%%%%%%%%
\clearpage

%%%%%%%%%%%%%%%%%%%%%%%%%%%%%%%%%%%%%%%%%%%%%%%%%%%%%%%%%%%%%%%%%%%%%%%%%%%%%%%%%%%%%%%%%%%%%%%

%% file: sections/table_multi_run.tex
\begin{table}[t]
\small
\setlength{\tabcolsep}{3pt}
\centering
  \begin{tabular}{cc|cc}
    \toprule
    % \hline
    \multirow{2}{*}{\textbf{Dimension}} &
    \multirow{2}{*}{\textbf{Model}} &
      \multicolumn{2}{c}{\textbf{Distribution Modeling}}  \\
        &&  R ({\textit{p-value}}) & {KL-Divergence}   \\
        
    \midrule     
        \multirow{11}{*}{\shortstack{\textbf{Creativity}}}
            & LLaVA-7B & 0.6363 (<.01) & 1.0721  \\
            & LLaVA-7B & 0.6548 (<.01) & 1.0127  \\
            & LLaVA-7B & 0.6392 (<.01) & 1.0464  \\
            &&&\\
            & LLaVA-13B   & 0.6548 (<.01) & 0.3734 \\
            & LLaVA-13B   & 0.6289 (<.01) & 0.3933 \\
            & LLaVA-13B   & 0.6257 (<.01) & 0.3706 \\
            &&&\\
            & InternVL2-8B   & 0.6363 (<.01) & 1.4903 \\
            & InternVL2-8B   & 0.6548 (<.01) & 1.4542 \\
            & InternVL2-8B   & 0.7153 (<.01) & 1.4658 \\
            
    \midrule    
        \multirow{11}{*}{\shortstack{\textbf{Originality}}}
            & LLaVA-7B &  0.7573 (<.01) & 0.5649 \\
            & LLaVA-7B &  0.7623 (<.01) & 0.5946 \\
            & LLaVA-7B &  0.7781 (<.01) & 0.5942 \\
            &&&\\
            & LLaVA-13B   & 0.7108 (<.01) & 0.2983 \\
            & LLaVA-13B   & 0.6757 (<.01) & 0.3297 \\
            & LLaVA-13B   & 0.7262 (<.01) & 0.3448 \\
            &&&\\
            & InternVL2-8B   & 0.8169 (<.01) & 0.5707 \\
            & InternVL2-8B   & 0.8044 (<.01) & 0.6203 \\
            & InternVL2-8B   & 0.8124 (<.01) & 0.5935 \\
            
    \midrule    
        \multirow{11}{*}{\shortstack{\textbf{Atypicality}}}
            & LLaVA-7B & 0.7282 (<.01) & 0.4226 \\
            & LLaVA-7B & 0.7470 (<.01) & 0.4043 \\
            & LLaVA-7B & 0.7061 (<.01) & 0.4301 \\ 
            &&&\\
            & LLaVA-13B   & 0.6332 (<.01) & 0.2617 \\
            & LLaVA-13B   & 0.5732 (<.01) & 0.2692 \\
            & LLaVA-13B   & 0.7126 (<.01) & 0.2367 \\
            &&&\\
            & InternVL2-8B   & 0.7838 (<.01) & 0.4041 \\
            & InternVL2-8B   & 0.7821 (<.01) & 0.4083 \\
            & InternVL2-8B   & 0.7838 (<.01) & 0.4041 \\
    \bottomrule
  \end{tabular}
  
  \caption{Results for multiple runs}
  \label{table:multi-run results}
\end{table}